\documentclass{article}

\usepackage{arxiv}

\usepackage[utf8]{inputenc} 
\usepackage[T1]{fontenc}    
\usepackage{hyperref}       
\usepackage{url}            
\usepackage{booktabs}       
\usepackage{amsfonts}       
\usepackage{nicefrac}       
\usepackage{microtype}      
\usepackage{lipsum}		
\usepackage{graphicx}
\usepackage{doi}
\usepackage[numbers]{natbib}
\usepackage{textcomp}
\usepackage{multirow}
\usepackage{makecell}
\usepackage{footnote}
\usepackage{xcolor}
\usepackage{algorithm}
\usepackage{algpseudocode}
\usepackage{rotating}
\usepackage{subcaption}
\usepackage{amsmath}
\newcommand{\ceil}[1]{\left\lceil #1 \right\rceil}

\usepackage{lineno,hyperref}
\usepackage{nicefrac}

\title{USEFUSE: Uniform Stride for Enhanced Performance in Fused Layer Architecture of Deep Neural Networks \thanks{Accepted for publication in the Journal of Systems Architecture on 11 May, 2025.}\thanks{This research was supported by Basic Science Research Program funded by the Ministry of Education through the National Research Foundation of Korea (NRF-2020R1I1A3063857). The EDA tool was supported by the IC Design Education Center (IDEC), Korea.}}

\date{} 					

\author{\href{https://orcid.org/0000-0002-1387-0879}{\includegraphics[scale=0.06]{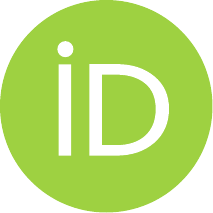}\hspace{1mm}Muhammad Sohail Ibrahim}\thanks{Muhammad Sohail Ibrahim and Muhammad Usman contributed equally to this work} \\
	Department of Mechanical Systems Engineering\\
	Kumoh National Institute of Technology\\
	Gumi-si, Republic of Korea \\
	\texttt{msohail@kumoh.ac.kr} \\
	\And
\href{https://orcid.org/0000-0002-3393-5211}{\includegraphics[scale=0.06]{orcid.pdf}\hspace{1mm}Muhammad Usman$^{\ddag \S}$} \\
	Faculty of Informatics and Data Science\\
	 University of Regensburg\\
	Regensburg, Germany \\
	\texttt{muhammad.usman@ur.de} \\
	\And
\href{https://orcid.org/0000-0002-5166-0629}{\includegraphics[scale=0.06]{orcid.pdf}\hspace{1mm}Jeong-A Lee}\thanks{Corresponding Authors}\\
	Department of Computer Engineering\\
	Chosun University\\
	Gwangju, Republic of Korea \\
	\texttt{jalee@chosun.ac.kr} \\
}

\renewcommand{\shorttitle}{USEFUSE}

\hypersetup{
}

\begin{document}
\maketitle

\begin{abstract}
Convolutional Neural Networks (CNNs) are crucial in various applications, but deploying them on resource-constrained edge devices poses challenges. This study presents the Sum-of-Products (SOP) units for convolution, which utilize low-latency left-to-right bit-serial arithmetic to minimize response time and enhance overall performance. The study proposes a methodology for fusing multiple convolution layers to reduce off-chip memory communication and increase the overall performance. An effective mechanism detects and skips inefficient convolutions after ReLU layers, minimizing power consumption without compromising accuracy. Additionally, efficient tile movement guarantees uniform access to the fusion pyramid. An analysis demonstrates the uniform stride strategy improves operational intensity. Two designs cater to varied demands: one focuses on minimal response time for mission-critical applications, and another focuses on resource-constrained devices with comparable latency. This approach notably reduced redundant computations, improving the efficiency of CNN deployment on edge devices.
\end{abstract}


\section{Introduction}\label{sec: Intro}
Deep neural network (DNN) is an artificial neural network comprised of several layers between input and output layers. They have been widely used in image recognition \cite{sun2020automatically}, semantic segmentation \cite{long2015fully}, medical imaging \cite{yoon2018efficient}, bioinformatics \cite{usman2022afp}, and signal processing \cite{chen2022citisen} etc. A class of DNN is the convolution neural networks (CNNs) which play a pivotal role in many applications such as computer vision, recognition, object detection, etc. This has been made possible due to the advancements in high performance computing technologies and the availability of cutting-edge compute resources. The use of CNNs with many layers has enabled the swift progress in a number of diverse application domains. CNN designs, inspired by the behavior of optic nerves in human brain, perform data processing in multiple layers of neurons to achieve human brain-like performance in image recognition.

There is a pressing need to execute complex neural networks in mission-critical applications with low latency demands. However, due to limited compute and storage resources, the implementation of neural networks on edge devices is limited \cite{chen2021dnnoff}. Various efforts have been made to reduce the complexity of neural networks at algorithm level, including compression by pruning, quantization, approximation, zero skipping, etc., at the expense of accuracy \cite{lin2023intermittent, oh2022non, danopoulos2022adapt}. Furthermore, several spatial architectures exploiting the effective data such as weight stationary \cite{yoo20151}, output stationary \cite{du2015shidiannao} and row stationary \cite{chen2016eyeriss}, have been proposed to accelerate the computation of neural networks.

In the context of network compression for resource-constrained hardware, serial processing is usually favored in DNN implementation where the models can have layer-specific input precision for either activation or weight (bit-serial with one operand in parallel) \cite{judd2016stripes, lee2018unpu} or both (bit-serial with both operands in serial)  as in \cite{sharma2018bit}. Parallel processing accounts for larger area, whereas serial approaches have longer computation time. Bit-serial designs however, require simpler circuitry and the adjustable precision makes them favorable for domain-specific hardware accelerators. Bit-serial designs also suffer from high latency and low throughput issues which are usually addressed by employing multiple instances of small serial circuits to deliver higher throughput \cite{judd2016stripes}.

In most modern CNNs, convolution accounts for more than $90\%$ of operations. Although the computation of the convolution operations is very simple, involving multiplication and addition, but due to the depth of such networks the computational complexity increases, subsequently raising the number of operations. It is found that nearly $85\%$ of the overall time in a CNN-based classification model is consumed by these multiplication and addition operations to perform convolution in the DaDianNao accelerator \cite{judd2016stripes}.
 
\begin{figure}[!ht]
    \centering
    \includegraphics[width=0.75\linewidth]{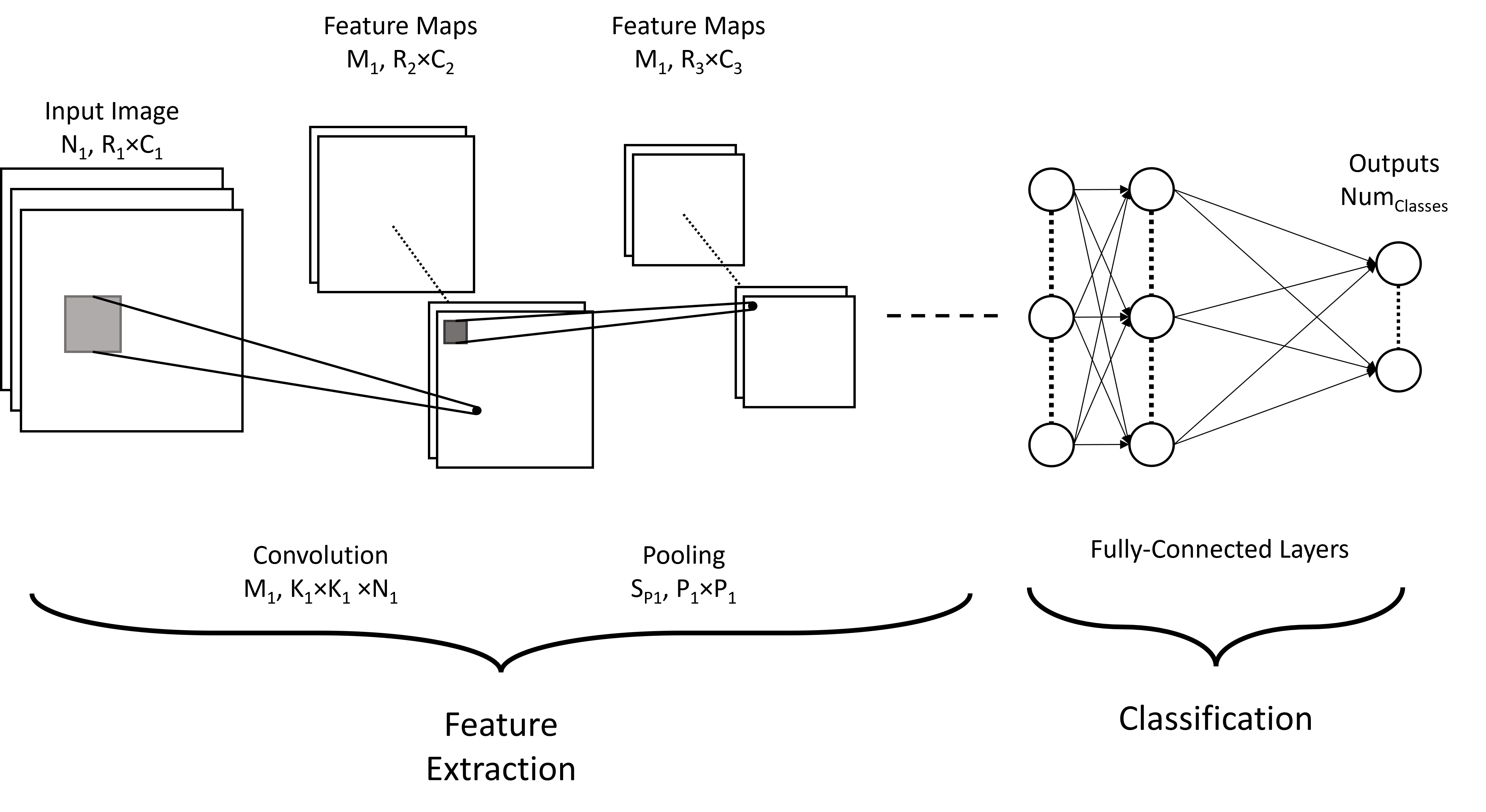}
    \caption{A general CNN architecture.}
    \label{fig:CNN-basic}
\end{figure}

A generic CNN architecture is illustrated in Fig.~\ref{fig:CNN-basic}, in which the feature extraction module includes stacks of convolution, activation, and pooling layers while the classification module contains stacks of fully connected layers. Activation functions to add non-linearity are placed either after convolution layer or pooling layer. One of the simplest and most commonly used activation function in CNNs is the rectified linear unit (ReLU), which returns same value if the input is positive and zero if the input is negative. It is reported that around $60\%$ of the convolutions produce negative output and therefore, their results are not used due to ReLU in the subsequent operations \cite{kim2021compreend}. Such convolutions have no effect on the output and are termed as ineffectual convolutions. In order to mitigate the computation of ineffectual convolutions, researchers have proposed methods to detect such negative activation and terminate them at an early stage, consequently resulting in energy and power savings \cite{akhlaghi2018snapea}. Several methods have been proposed to detect these negative results at an early stage and terminate them to reduce energy consumption \cite{lee2018compend, chen2019comprrae, kim2021compreend}. However, it adds an overhead to the model.
This overhead can be alleviated by utilizing an unconventional arithmetic known as \emph{online arithmetic} \cite{ercegovac2004digital}, where all the computations are performed from left-to-right in most significant digit first (MSDF) manner.

Moreover, the information in the CNNs flows sequentially, i.e., the results from one layer are utilized by subsequent layers. This inter-layer connectivity forms the backbone of CNNs, allowing them to progressively uncover intricate features as the data travels deeper into the network. Conventional CNN accelerator implementations perform the network computation in a layer-by-layer manner where the data is repeatedly transferred between the processing units and memory. Such communication burden increases rapidly with the increasing depth of the network.

In this avenue, exploring the dataflow across the CNN layers can help in the reduction of memory traffic. Instead of layer-by-layer computation in a CNN, the CNN layers can be \textit{fused} together to reduce the communication between memory and the compute engines \cite{alwani2016fused}. In such architectures, it is well-defined in the literature that each pixel or receptive field in the activation of a particular hidden layer is dependent on a region in the input activation of the initial layer of the network. Using a layer fusion strategy can help in merging the operations of various subsequent layers, in-turn reducing the off-chip memory access substantially.

In this work, we develop a CNN evaluation scheme by fusing the convolution layers of the network to reduce the off-chip memory traffic due to the intermediate data. The proposed approach reduces the number of duplicate computations by an efficient uniform fusion pyramid movement scheme aided by a uniform stride for each level of the fusion design. The goal is to select the smallest possible tile sizes for each layer in the fusion design, maintaining a uniform tile movement while ensuring minimum overlap between the adjoining tiles. The contributions of this study can be summarized as follows:

\begin{itemize}
    \item Design of SoP units for convolution using low-latency left-to-right bit-serial arithmetic based computation units to minimize the response time.
    \item A methodology to fuse several layers of convolution neural network using the proposed SoP computation units to decrease the off-chip memory communication.
    \item Mechanism to early detect and skip the computation of convolutions which are ineffectual after ReLU layers without any loss in accuracy during inference to minimize the power consumption.
    \item A methodology for tile movement to ensure efficient data access and uniform movement of fusion pyramid.
    \item An analysis depicting that the proposed uniform stride strategy improves the operational intensity irrespective of the dataflow of the underlying computation units.
    \item Additionally, we propose two alternate designs possessing the aforementioned properties; 1) a design aimed at minimal response time for mission critical applications, 2) a design suitable for resource constrained devices with comparable latency as the contemporary approaches.
\end{itemize}

The rest of the paper is organized as: a comprehensive review of relevant literature is discussed in Section~\ref{sec:2}, the proposed CNN evaluation scheme is presented in the subsequent Section~\ref{sec:3}. The experimental results and relevant discussion is presented in Section~\ref{sec:4} followed by the conclusion of the study in Section~\ref{sec: Conclusion}.

\section{Related Work}\label{sec:2}

This section will review existing bit-serial architectures for CNN computation, followed by a discussion on fused-layer architectures. Finally, methods for early detection and termination of negative computations to enhance energy efficiency will be explored.
\subsection{Bit-Serial Accelerators}
Over the past decade, researchers have addressed various challenges in CNN acceleration, such as unnecessary computations and the need for variable precision across CNN layers \cite{judd2018proteus, shin2017fixed}. These challenges can lead to increased energy and resource demands in accelerator designs. Stripes \cite{judd2016stripes}, a leading CNN acceleration design, employs bit-serial arithmetic compute units to exploit variable precision, thereby speeding up CNN inference.  

The primary goal of bit-serial arithmetic designs is to reduce unnecessary computations and create energy-efficient accelerators. In this direction, Bitlet \cite{lu2021distilling} proposed a bit-interleaving architecture that leverages bit-level sparsity and variable precision to accelerate DNN inference. A mixed-precision CNN accelerator is presented in \cite{latotzke2022design}, achieving high throughput with minimal accuracy loss by quantizing inputs and weights. Similarly, T-DLA \cite{chen2019t} uses 2-bit quantized weights for performance improvements. TALIPOT \cite{karadeniz2021talipot} enhances energy efficiency using hybrid number representations in most significant bit first (MSBF) arithmetic units, allowing early operations in subsequent layers without waiting for complete computation results. Other variable precision, bit-serial computation-based designs include implementations such as \cite{lee2018unpu,  liu2020precision, albericio2017bitpragmatic}.

Bit-serial designs provide advantages such as reduced memory bandwidth requirements and the ability to leverage variable precision across different DNN layers. However, these designs face drawbacks, including higher latency, lower throughput, and reduced performance compared to conventional bit-parallel architectures. Additionally, in bit-serial designs, the accumulation operation is hindered by carry propagation, which significantly increases cycle time and lowers the operating frequency of the processing units \cite{al2020towards}.

\subsection{Fused-Layer Accelerators}

Conventional CNN accelerator designs focus on iterative layer computations, generating large amounts of intermediate data. Depending on the design and tile size, this data can be intra-layer or inter-layer, requiring off-chip memory storage and retrieval for subsequent operations. As CNN models grow deeper, memory traffic increases. To address this, a novel accelerator design that *fuses* multiple CNN layers was introduced in \cite{alwani2016fused}, reducing intermediate memory traffic by directly feeding data between adjacent compute units, minimizing off-chip memory use by up to $95\%$ for models like VGGNet-E.

Fused-layer architectures can also take advantage of variable precision requirements in different layers, improving efficiency. Bit-Fusion \cite{sharma2018bit} introduced a flexible architecture where bit-level processing elements dynamically fuse to match precision needs, increasing speed and energy efficiency without accuracy loss. Efficient computation scheduling is critical for fused-layer designs, as highlighted by ConvFusion \cite{waeijen2021convfusion}, which proposed a cost model for scheduling computation and memory communication, optimizing tiling, loop reordering, data reuse, layer fusion, and convolution execution schemes. Other layer-fusion-based designs include DeepThings \cite{zhao2018deepthings}, TGPA \cite{wei2018tgpa}, and further approaches explored in \cite{xiao2017exploring, li2022fused, zhou2022accelerating}.

The data flow between computation units and external memory presents significant design challenges and increased energy consumption due to the large volume of data generated during CNN convolution operations \cite{alwani2016fused}. Fused-layer architectures attempt to mitigate this by reusing intermediate data, but exploring the design space for data scheduling, loop tiling, and loop reordering remains challenging.

Olympus \cite{cai2021olympus} addresses memory access traffic by optimizing both intra-layer and inter-layer data reuse. It employs a memory-oriented network scheduling technique to reduce memory traffic and enhance energy efficiency in DNN processors. Other strategies for minimizing memory access and exploring accelerator design space include \cite{tewari2021minimizing, ahmad2020superslash, tewari2020bus, kang2021multi}.

Despite the benefits of fused-layer dataflow, certain limitations remain. Many fused-layer designs overlook the stride of the fusion tile, which determines how the tile moves after computation. Incorrect stride determination can lead to excessive duplicate data being reused or recomputed, requiring large buffers or on-chip storage for intermediate data and causing significant under-utilization of compute resources. Storing intermediate data has been shown to be more energy-efficient than recomputation \cite{alwani2016fused}. The need for large data buffers arises due to two factors: (1) ineffective computation of tile stride and (2) the use of conventional arithmetic units that fail to process the generated data immediately.

\subsection{Early Negative Detection Techniques}

Rectified Linear Activation (ReLU) is a popular activation function in neural networks, which sets negative values to zero while keeping positive values unchanged. \textcolor{black}{With the advancement in deep learning architectures, various derivatives of the ReLU activation functions have been proposed, such as PReLU \cite{zhang2023practical}, LeakyReLU \cite{li2024image}, etc., while many recent architectures still rely on the ReLU activation \cite{zhang2021plug, chen2024artadapter, bhattad2024stylitgan, tafasca2024sharingan}. Additionally, recent research suggests that ReLU can serve as a viable alternative to softmax, offering advantages in computation efficiency and parallelization. For instance, \cite{shen2023study} proposed a ReLU based self-attention and feed-forward network to replace softmax in transformer models, showing that ReLU improves scalability by efficiently handling a large number of memory slots. Similarly, \cite{wortsman2023replacing} replaced softmax with ReLU in vision transformers and demonstrated that the ReLU-based attention achieves comparable performance to softmax-based attention in terms of scaling behavior while enabling better parallelization over the sequence length dimension, reducing the need for gather operations. These findings indicate that ReLU is not only relevant but also increasingly explored as a substitute for more complex functions in deep learning architectures.} The introduction of ReLU in the network architecture facilitates faster convergence and helps address the vanishing gradient problem. However, ReLU introduces the issue of ineffectual convolutions, where a significant portion of a convolution layer's output consists of zero activations after applying ReLU. These zeros are propagated through the network without contributing to the final output, leading to wasted computational resources. This inefficiency consumes memory bandwidth, energy, and processing cycles, ultimately slowing down inference and increasing energy consumption.

As mentioned earlier, while many DNN acceleration techniques focus on designing fast and energy-efficient computation units, fewer approaches address the early termination of convolution operations due to ReLU activations. SnaPEA \cite{akhlaghi2018snapea} introduced an early negative prediction scheme with two modes to address this: 1) Exact Mode: A single-bit sign check is performed iteratively on the sum of partial products, and computation stops as soon as the sum falls below zero, and 2) Predictive Mode: The partial sum is compared to a threshold, and computation is terminated if it drops below this threshold. This mode is faster but slightly reduces accuracy. Other methods aimed at early termination of convolution operations include CompreEND \cite{kim2021compreend}, TermiNETor \cite{mallappa2022terminetor}, CompRRAE \cite{chen2019comprrae}, CompEND \cite{lee2018compend}, BitSET \cite{pan2023bitset}, and \cite{asadikouhanjani2020novel}.

Left-to-right or MSDF arithmetic operations can significantly enhance the early detection of negative activations. Shuvo et al. \cite{shuvo2020msb} proposed a novel circuit implementation for convolution that allows for early detection of negative results, enabling the subsequent termination of related operations. However, existing methods for early detection of negative activations often rely on digit encoding schemes, threshold-based predictions, or complex circuitry, which can result in erroneous decisions or increased overhead.

\section{Materials and Methods}\label{sec:3}
To address the limitations of the existing works, we propose to utilize digit serial left-to-right arithmetic-based computation units, terminating the computation of ineffective convolutions at an early stage, and minimize the communication between memory and compute units by fusing several successive convolution layers. The details of which have been explained in the ensuing subsections.
\subsection{Online Arithmetic}

In online arithmetic, computations proceed digit-by-digit, from the most to the least significant position, for both inputs and outputs. Algorithms require $(j + \delta)$ input digits to compute the $j^{th}$ digit of the result, where $\delta$ is the online delay, typically a small integer (1-4) depending on the operation. This method employs a redundant number system to generate the most significant digits first, making the cycle time independent of the working precision. \textcolor{black}{Online algorithms involve recurrence relations where residuals are iteratively fed back into computations. The residual part from intermediate calculations contributes to generating subsequent output digits efficiently.}

Online arithmetic enables the overlap of dependent operations, as the subsequent unit can begin computation once the most significant digit (MSD) of the preceding unit is available. In contrast, conventional digit-serial arithmetic requires all digits before starting. Although overlapping is possible in conventional systems if all operations use either MSDF or least significant digit-first (LSDF) modes, issues arise when combining MSDF (e.g., division) with LSDF (e.g., multiplication). Since online arithmetic consistently uses MSDF, it supports seamless overlapping of dependent operations. \textcolor{black}{ In conventional arithmetic, the subsequent unit can only begin computation if the output of the preceding unit is generated bit-by-bit and the subsequent unit also accepts input bit-by-bit. Otherwise, if it requires a parallel input, it must wait until the entire output is available. Online arithmetic, however, takes input serially and produces results serially, enabling a technique called \textit{computation while communication}, where processing and data transfer occur simultaneously, reducing latency and improving efficiency.} 

In parallel or pipelined systems where full-precision communication between modules is not feasible, online arithmetic excels due to its reduced bandwidth needs. This is particularly advantageous in signal processing applications where full-precision output is unnecessary. For instance, in multiplying two $N$-bit operands to generate a $2N$-bit result, often only the most significant half is required, as in many DSP applications. Conventional multipliers produce output starting from the least significant bits, discarding the lower half and wasting resources. In contrast, online arithmetic generates output digit-by-digit from the most significant side, allowing computation to stop once the desired precision is reached.

The computation from the most significant digit (MSD) to the least significant digit (LSD) relies on generating output based on partial information about the input operands. This flexibility is achieved by introducing redundancy in the input and output operands, which is why a redundant number representation system is used in online arithmetic. Typically, a signed digit (SD) redundant number system is employed, where numbers are represented in radix \( r \) form, and each signed digit belongs to the set \(\{-a, \ldots, -1, 0, 1, \ldots, a\}\) with the condition \(\frac{r}{2} \leq a < r\). In this work, we utilize a symmetric radix-2 digit set with \(\{-1, 0, 1\}\).

\subsubsection{Online Multiplier and Adder Overview}

The fundamental component of the accelerator is the window processing unit (WPU), which serves as the core for computing convolutions. The WPU is composed of online multipliers and reduction trees based on online adders. In the online serial-parallel multiplier, one operand is fed in serially in a most significant digit first (MSDF) manner, while the other operand is a constant available in parallel at implementation time. A radix-2 serial-parallel online multiplier has an online delay of 2, and its selection function requires 2 fractional bits and 1 integer bit for output digit selection. Methods for developing online algorithms and derivations are discussed in \cite{ercegovac2004digital}. The online multiplication algorithm generally consists of two steps: 1) Initialization, during which \(\delta\) input digits (in serial) are collected without generating any output, resulting in an execution length equal to the online delay (\(\delta\)); and 2) Recurrence, which runs for \(n\) iterations, where \(n\) is the input precision, producing one output digit in each iteration. A pseudo-code for the online serial-parallel multiplication algorithm is presented in Algorithm~\ref{alg:algorithm_SP}.
   \begin{algorithm}
		  	\begin{algorithmic}[1]
		  	\State {Initialize:\newline $x[-2]=w[-2] = 0$}
            \For{j=$-2,-1$}
                \State{$v[\jmath]=2 w[\jmath]+\left(x_{j+2} \cdot Y] \right) 2^{-2}$}
                \State{$w[\jmath+1] \leftarrow v[j]$}
            \EndFor \newline
             \State{Recurrence:}
             \For{$j=0 \ldots n+\delta$}

                \State{$v[\jmath]=2 w[\jmath]+\left(x_{j+2} \cdot Y] \right) 2^{-2}$}
                \State{$z_{j+1}=SELM(\widehat{v[j]})$}
                \State{$w[j+1] \leftarrow v[j]-z_{j+1}$}
                \State{$Z_{\text {out}} \leftarrow z_{j+1}$}
             \EndFor \newline
            
\end{algorithmic}
\caption{Serial-Parallel Online Multiplication}
\label{alg:algorithm_SP}
  \end{algorithm}

Here, $x$ and $Y$ are the bit-serial and parallel inputs, respectively, and $z$ is the serial MSDF output. The residual registers to store the temporary results are denoted by $\omega$ and $v$. At any $j^{th}$ iteration, the serial output digit (input digit) is represented by  $z_j$ ($x_j$), where $z_j = SUB(z^+, z^-)$, such that the subtraction of the two bits represents the value of the digit. $SELM(.)$ is the output selection module/function that selects an output from a look-up table on the basis of a few most significant ($t$) bits of the residual. 

Serial online addition involves full adders and registers to add two redundant numbers in a MSDF manner. A detailed description of the online adder and its relevant derivations can be found in Ercegovac and Lang \cite{ercegovac2004digital}. Additionally, this reference provides the design and methodology for the online serial-serial multiplier, where both input operands are supplied as serial inputs. In this work, we utilize the online serial-parallel multiplier proposed in our previous research \cite{usman2023low}. This online serial-parallel multiplier is employed to design the processing units in the proposed USEFUSE accelerator, with further details and derivations available in \cite{usman2023low}.

\subsection{Early Termination of Negative Computations}
Most CNN accelerator designs concentrate on efficiently generating the SOP for the activation layer (ReLU). However, few studies have investigated the early detection of negative values in the SOP, which presents a significant challenge in accelerators based on conventional arithmetic. For instance, conventional bit-serial multipliers take the multiplicand in parallel while processing the multiplier serially. In each iteration, a partial product is generated and stored in a register, then shifted into the appropriate position before being added to other partial products to compute the final result. This process typically involves a series of adders for reduction. A second level of reduction is necessary to add \( k \times k \) products to yield the output pixel, along with an additional level of reduction for summing multiple input channels. With conventional bit-serial multipliers, the most significant bit and the polarity of the result cannot be determined until all partial products have been generated and added to the previous partial sums.

The challenge of early detection and termination of negative activations can be addressed by the intrinsic ability of online arithmetic to generate output digits in an MSDF manner. The proposed design supports the termination of negative activation computation in $p<\mathcal{N}$ cycles, where $\mathcal{N}$ is the number of cycles to compute complete result. This is done by observing the output digits. The process of detecting the negative activations and subsequently terminating the relevant computation is summarized in Algorithm \ref{alg:ENT}.

\begin{algorithm}
\caption{Early detection and termination of negative activations}\label{alg:ENT}
\begin{algorithmic}
\State{$z^{+}_j,\ z^{-}_j \ $ bits} 
\For{$j = 1 \ to \ \mathcal{N}$}
    
        \State $z^{+}[j] \gets z^{+}[j]\ ^\frown z^{+}_{j}$\ 
        \State $z^{-}[j] \gets  z^{-}[j]\ ^\frown z^{-}_{j}$\
        \If{$z^{+}[j] \ < \ z^{-}[j]$}
         \newline   Terminate
        
        \Else
         \newline   Continue
        \EndIf
      \EndFor    
\end{algorithmic}

\end{algorithm}

\noindent The proposed early negative detection unit (END-U) is equipped with registers to store \( z^{+}_j \) and \( z^{-}_j \) bits, which represent the positive and negative output bits of the sum of products (SOP) in redundant form. During each iteration, new bits are appended to their respective registers. As soon as the value of \( z^{+}[j] \) falls below the value of \( z^{-}[j] \), a termination signal is generated, resulting in the cessation of the SOP computation. The END-U is integrated into each processing unit, as described in Section \ref{sec:accelerator}.

\subsection{Proposed Layer Fusion Method}

\begin{figure}[!ht]
    \centering
    \includegraphics[width=0.65\linewidth]{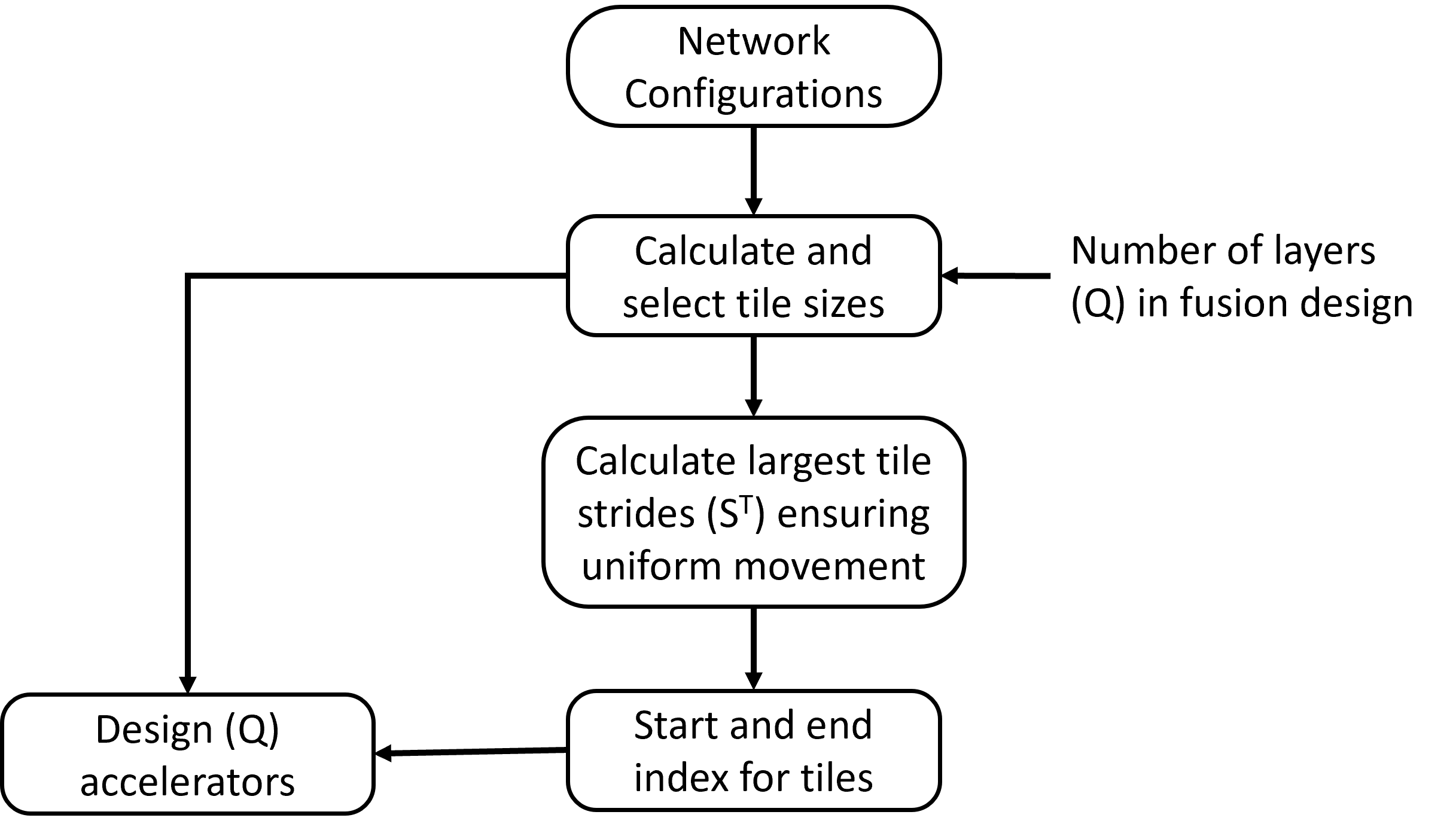}
    \caption{Proposed layer fusion accelerator design pipeline}
    \label{fig:flowchart}
\end{figure}

This section outlines the proposed layer fusion method and its components, including the calculation and selection of tile sizes and the calculation of the utile stride for tiles. A comprehensive description of the proposed design flow is presented in Fig.~\ref{fig:flowchart}. 
The design flow begins by taking the CNN network configurations and the number of \( Q \) convolution layers intended for the fusion design, followed by the calculation of tile sizes for each layer. This is followed by calculating the utile stride to ensure uniform tile movement across the respective layers in the fusion design. Next, the start and end indices of the feature maps intended for each layer are determined. The information accumulated throughout this process is then utilized to design accelerators for each layer in the fusion design, with detailed descriptions of these processes provided later in this section.

\subsubsection{Overview}
\begin{figure}
    \centering
    \includegraphics[width=0.45\linewidth]{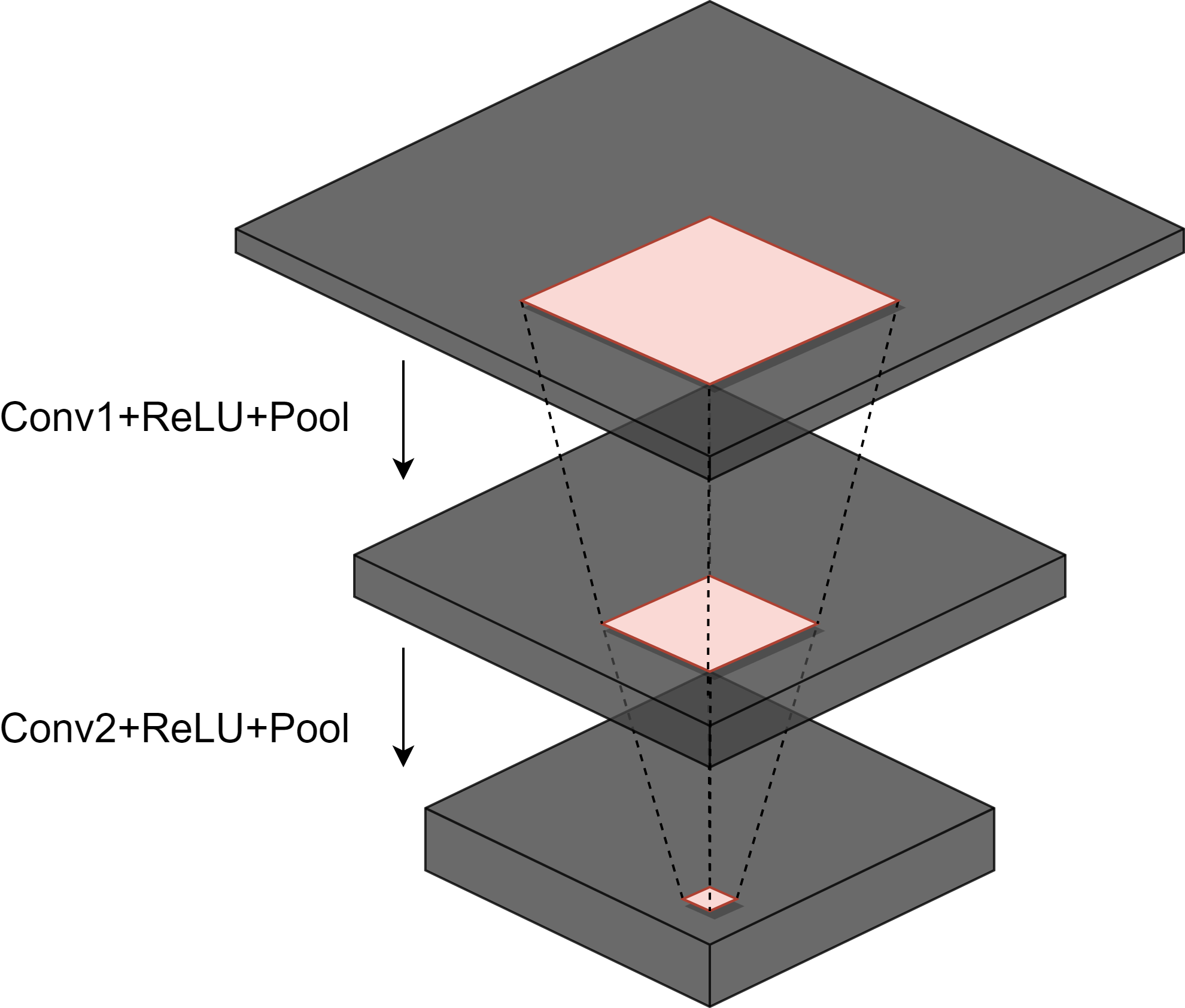}
    \caption{General layer fusion scheme}
    \label{fig:fused}
\end{figure}

The proposed design follows a layer fusion scheme as depicted in Fig.~\ref{fig:fused}, where a particular region, referred as \textit{Tile}, is selected by tracking the output activation (or a region) of the final layer of the fusion pyramid to the first layer. The dimensions of the tile depends on the CNN architecture as well as the dimensions of the intended region of the output feature map. 

The pyramid dimensions are calculated by selecting a suitable region of the output feature map and the tile dimension of its preceding layer according to relation \eqref{eq1}, presented in \cite{alwani2016fused}.

\begin{equation} \label{eq1}
    D_{l} = (D_{o} - 1) \times S_{l} + K_{l}
\end{equation}

where $D_{l}$ is the dimension of the layer preceding the output layer of the fusion pyramid, $D_{o}$ is the dimension of the selected region of the output feature map, $S_{l}$ and $K_{l}$ are the stride and kernel size of the layer preceding to the output layer, respectively. This procedure is done from the final layer until the first layer of the fusion architecture to obtain the tile sizes of the respective layers in the fusion pyramid.
Consider an example of a simple CNN such as LeNet-5 whose first two convolution layers are to be fused. 
Each convolution layer is followed by a sub-sampling layer, like Maxpooling. In a fusion of two convolution layers, $R = C = 1$ output pixels from the second sub-sampling layer serve as input to the third layer. To determine tile dimensions in the fusion pyramid, Eq.~\eqref{eq1} applies to both convolution and sub-sampling layers. For instance, the input to the third layer follows a Maxpooling operation ($MPL2$) on a $2\times2$ output from the second convolution layer ($CL2$), which operates on a $6\times6$ input. Tracing back, $MPL1$ requires a $12\times12$ input to produce this, derived from a $16\times16$ input to $CL1$. The generation of neighboring pixels at the same level requires a separate, overlapping pyramid computation. The starting index for each layer in this process, known as the \textit{tile stride}, differs from the convolution stride. Determining this \textit{tile stride} is crucial for two reasons: 1) It ensures the fusion pyramid covers the entire input feature map without skipping pixels, generating all necessary output activations. 2) It guarantees consistent execution rounds at every pyramid level, removing the need for synchronization after each round \cite{wei2018tgpa}. 

\textcolor{black}{Furthermore, in most CNN models, the feature map dimensions are downsampled along the depth of the network while the number of filters increase. The proposed scheme ensures reduction in memory traffic in earlier as well as later convolution layers. This is due to the reason that the proposed design incorporates input and output channel tiling \cite{zhang2015optimizing}. This means that the filters are loaded into the kernel buffers only once, while the input feature map sections are loaded into the input buffers as the fusion tile moves across the input feature map.}

To this end, we propose an algorithm in section \ref{sec:alg} for the calculation of the \textit{tile stride} to ensure a uniform movement of the fusion pyramid for various output region configurations including the tile dimensions for each layer in the fusion pyramid. It is also worth noting that this work focuses on the assumption that the tile at each pyramid level is square-shaped, which is most commonly used.

\subsubsection{Algorithm} \label{sec:alg}

The pseudocode presented in algorithm \ref{alg_tile_size} depicts a simple framework for calculating the fusion tile sizes of any network using Eq.~\eqref{eq1}. It takes the name of the network and the number of layers ($Q$) intended for fusion as its input and returns the fused-layer tile sizes for all possible squared output dimensions in the output feature map of the final layer in the fusion pyramid. It ensures that the tile size $H$ for each layer in the fusion design is bounded by the size of the input feature map ($IFM$) of the respective layer. The $For$ loop iterates over the various squared dimensions of the output feature map ($R_Q$) of the fused-layer design and results in an ($R_Q \times Q$) matrix consisting of tile sizes ($H_Q, H_{Q-1}, \dots, H_1$) for each layer in the fusion pyramid. This results in all possible fused-layer tile configurations considering that the tile sizes and respective outputs and inputs of each layer are square.

\begin{algorithm} 
\caption{Calculation for the Fusion Pyramid Tile Sizes}\label{alg_tile_size}
\begin{algorithmic}[1]
\Require Network, Number of Layers $Q$ 
\Ensure  $H \leq IFM$

\For{(i in $R_Q$)}
\For{($j=Q$, $j\geq 1$, $j$--)}
    \State{$H_{(i,j)} = (i - 1)\times S_j + K_j$}
    \EndFor
\EndFor

\State{Return $\mathbf{H} \in \mathbb{R}^{R_Q \times Q}$}
\end{algorithmic}
\end{algorithm}

Algorithm \ref{alg_tile_size} results in a relatively large design-space which can be narrowed down further by determining the appropriate stride for each tile in the fusion pyramid. The algorithm determines the number of movements $\alpha$ that a particular tile should take under various tile stride $S^T$ values. The $S^T$ values are calculated using the condition that $\alpha$ can only be an integer. Each value of $S^T$ dictates the amount of overlap between the adjoining tiles in a layer in the fusion pyramid. In order to ensure the least amount of overlap, an $S^T$ value of $H-K+S$ can be selected. Although this selection ensures the least amount of overlap as well as the least number of $\alpha$, but it can result in a different number of movements at different levels of the pyramid. For instance, in the previous example of LeNet-5, the tile size for $CL1$ and $CL2$ were selected to be $16 \times 16$ and $6 \times 6$ respectively. The tile stride for $CL1$ and $CL2$ will result in $S^T_1 = 16-5+1 = 12$ and $S^T_2 = 6-5+1 = 2$, respectively. The value of $S^T_2$ shows that the tile representing $CL2$ results in $\alpha_2 = 5$, while the value of $S^T_1$ results in a non-integer value of $\alpha_1 = \nicefrac{7}{3}$ which has to be ruled-out. Also, the movement parameters for $CL1$ and $CL2$ do not agree, resulting in an asymmetric movement of different tiles in the fusion pyramid. This can lead to a number of issues; 1) requiring some synchronization delay between the execution of tiles caused by the stall cycles inserted between the execution of adjoining tiles, 2) increased latency due to one tile being executed several times more due to repeated computations compared to others in-turn decreasing the overall operating frequency of the design, and 3) the mismatch in synchronization may require for some intermediate data to be shuttled back to the memory in case of limited buffer space.  

\begin{algorithm} 

\caption{Calculation for the Tile Stride}\label{alg_tile_stride}
\begin{algorithmic}[1]
\Require $\mathbf{H}$ $\in$ $\mathbb{R}^{R_Q \times Q}$ 
\For{$i = 1$, $i \leq R_Q$, $i$++}
\For{$j = 1$, $j \leq Q$, $j$++}
\For{$p = 1$, $p \leq H_j$, $p$++}
    \State{$\alpha_{(i,j,p)}$ = $\frac{IFM_j - H_j}{p} + 1$}
        \If{$\alpha_{(i,j,p)}$ $\in$ $\mathbb{Z}$}
        \State{$\mathbf{\alpha_{i,j}}$ $\gets$ $\alpha_{(i,j,p)}$}
        \State{$S^{T}_{i,j}$ $\gets$ $p$}
\EndIf
\EndFor
\EndFor
\EndFor
\State{Return $\mathbf{\alpha, S^T} \in \mathbb{R}^{R_Q \times Q}$}
\end{algorithmic}
\end{algorithm}

After calculating the $S^T$ and $\alpha$ parameters for the fusion tile size $\mathbb{H} \in \mathbf{H}$ of choice, the values of $S^T$ resulting in the same $\alpha$ parameter values for each layer in the fusion pyramid can be evaluated and the corresponding $S^T$ values for each layer can be obtained. The appropriate $S^T$ values for each layer resulting in a synchronized fusion pyramid movement can simply be obtained by analyzing that the candidates for $S^T$ do not result in skipping the computation of some regions in any layer. Among these $S^T$ candidates, the maximum values for $S^T$ for each layer is carefully selected after satisfying the condition stated earlier. Such $S^T$ values ensure a uniform movement of each tile in the fusion pyramid, thereby addressing the three problems stated earlier. 

\subsection{Accelerator Designs} \label{sec:accelerator}
In order to show the efficacy of the proposed technique, we present two distinct approaches to the accelerator design. In one of the configurations, we aim to minimize the latency of the computations by exploiting the spatial parallelism in convolution operation at the cost of area. However, we show that conventional arithmetic-based design with the same configuration does not match the latency and the performance provided by the use of online arithmetic-based components. 
Additionally, an alternative, more pragmatic design is introduced, which performs convolution in a temporal manner and efficiently utilizes limited computational resources. 
\begin{figure}[!ht]
    \centering
    \includegraphics[width=0.45\linewidth]{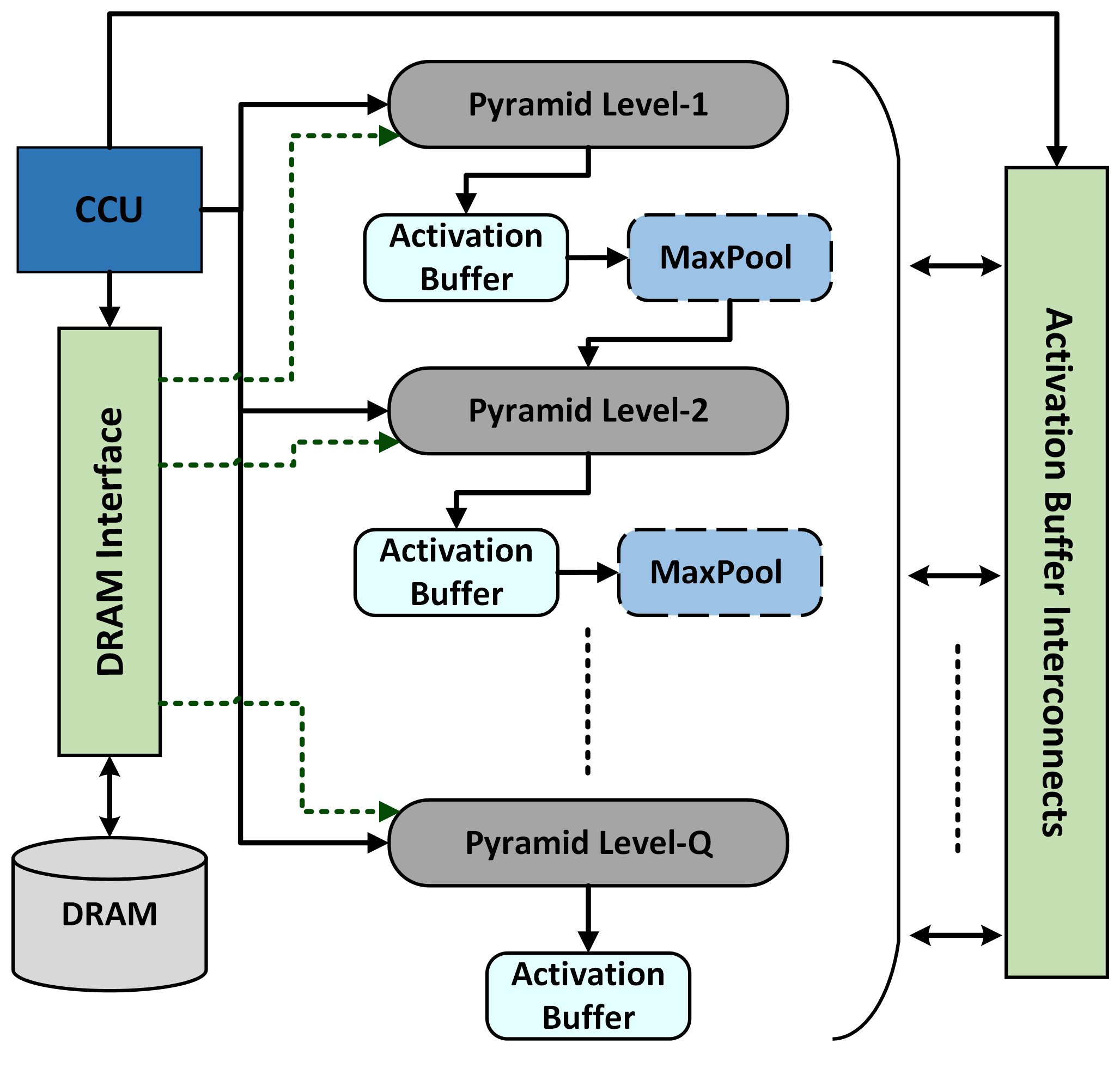}
    \caption{{Overall Architecture. The solid black arrows represent the input, output, and control connections, while the dotted green arrows represent the filter/weight data.}}
    \label{fig:arch}
\end{figure}

\textcolor{black}{Both the aforementioned designs have similar general overall accelerator architecture. The overall architecture of the proposed accelerator is presented in Fig.~\ref{fig:arch}. Depending on the number of convolution layers in a CNN model, there can be many pyramid levels in the proposed fused-layer design. Each pyramid level represents a tile in a particular convolution layer of the CNN model. The depth $Q$ of the fusion determines the number of levels in the pyramid. The selection of the depth $Q$ of the fusion pyramid can also help in optimizing the performance of the layer fusion acceleration designs. However, this work focuses primarily on the calculation and selection of tile sizes and uniform stride, the use of online arithmetic-based compute units, early negative detection, etc. The optimization related to the selection of the number of layers ($Q$) in the fusion pyramid is considered as future work. In the current work, the parameter $Q$ is selected as referenced in literature \cite{alwani2016fused, wei2018tgpa}. Each pyramid level is followed by an activation buffer and an optional pooling block. The activation buffer block offers an on-chip buffer storage for the output features of the previous pyramid level. It is also noteworthy that for implementation on FPGAs, a large value of $Q$ cannot be feasible due to the limited resources. However, we performed an experiment with 4 convolution layers of VGG-16 CNN and the experimental results show that with enough hardware resources, the proposed technique can be utilized for larger $Q$ values.}

\textcolor{black}{The proposed technique for the tile stride selection ensures uniform tile movement across the different pyramid levels. However, it leads to a slightly larger area of overlap regions within the fusion pyramid feature maps compared to the ($H-K+S$) region, which ensures minimum overlap. However, it is noteworthy that the proposed tile stride calculation technique not only ensures uniform movement across the pyramid levels but also keeps the number of pyramid movement $\alpha$ to a minimum. This ensures that the overlap region does not increase drastically (ensured by the larger tile stride values). Furthermore, the overlapped output pixels of a pyramid level are stored in the output buffers to be reused by the subsequent level in the fusion pyramid as the fusion tile moves across the input feature map for its computation. This means that the proposed USEFUSE design performs output pixel reuse instead of recompute as suggested in \cite{alwani2016fused}.}

\subsubsection{Design Strategy-1 (DS-1) - Spatial Design}
\begin{figure}[!ht]
    \centering
    \includegraphics[width=0.8\linewidth]{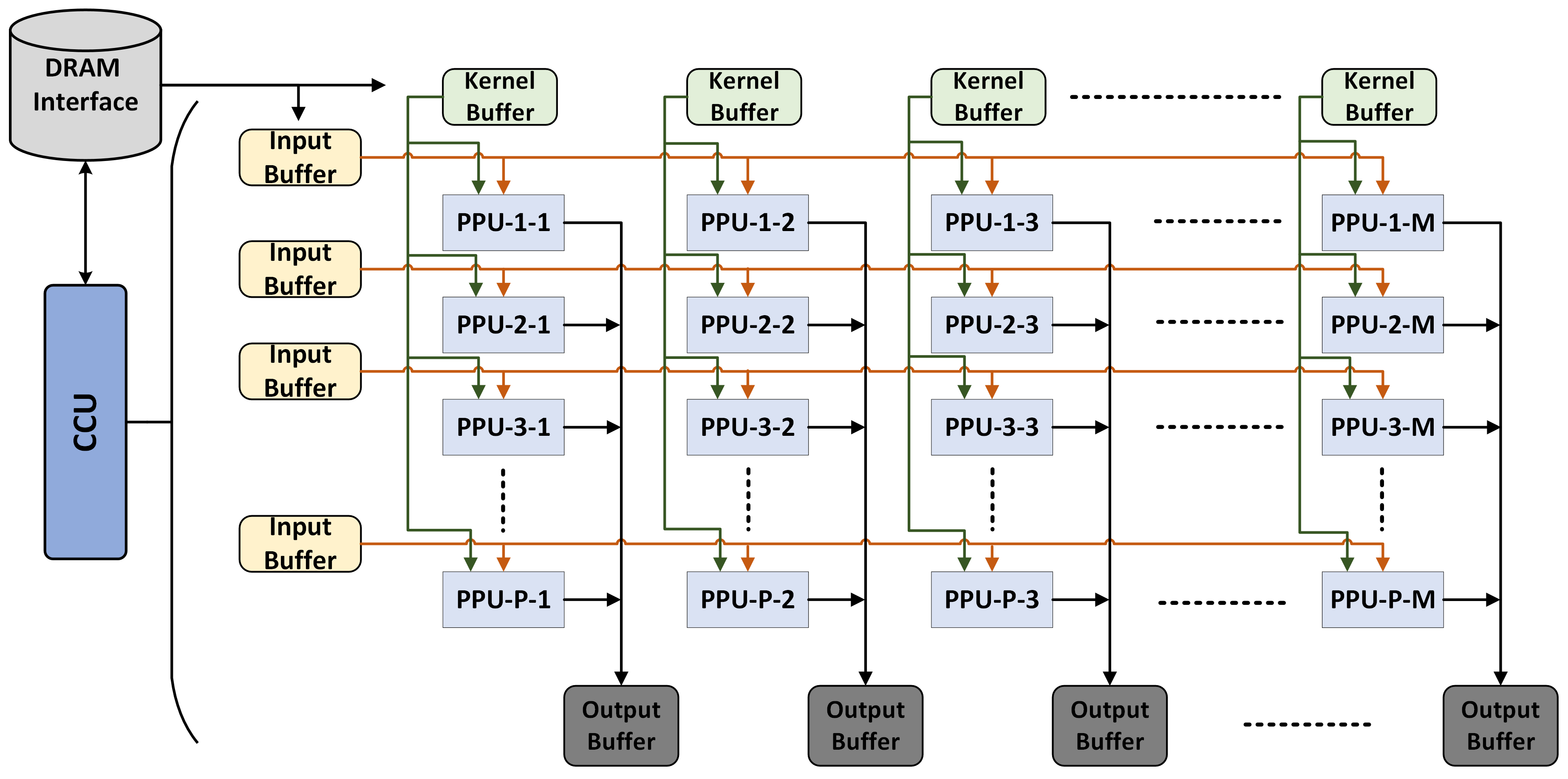}
    \caption{Tile/Pyramid Level Design}
    \label{fig:tile}
\end{figure}

Each pyramid level in Fig.~\ref{fig:arch} represents an accelerator of the respective convolution layer in the fusion pyramid. A general architecture of the accelerator is presented in Fig.~\ref{fig:tile}. It is composed of $P=R\times C$ rows and $M$ columns, where $R\times C$ is the dimension of the output of a tile $\mathbb{H}$, and $M$ is the number of output feature maps of the respective convolution layer. Each \textit{input buffer} broadcasts the input data of a unique convolution window to its corresponding row of pixel processing units (PPUs). The \textit{kernel buffers} broadcast the convolution filter to each PPU in a column. The \textit{output buffers} collect the pixels from each PPU.  

Each row of the array computes a unique $K\times K \times N$ window of the input feature map of the corresponding layer within the fusion pyramid. The $K\times K \times N$ input to each row, represented by the \textit{input buffer}, is broadcasted to each PPU in the corresponding row. The array architecture presented in Fig.~\ref{fig:tile} also shows that the accelerator array supports output tiling \cite{zhang2015optimizing} ($t_{m}=M$) as the number of columns in the array represent the number of output feature maps.
It consists of an array of pixel processing units (PPU), where each column of the PPU computes a distinct output feature map (OFM). The filter corresponding to each OFM is broadcasted to every PPU in the respective column.

\begin{figure}[!ht]
    \centering
    \includegraphics[width=0.55\linewidth]{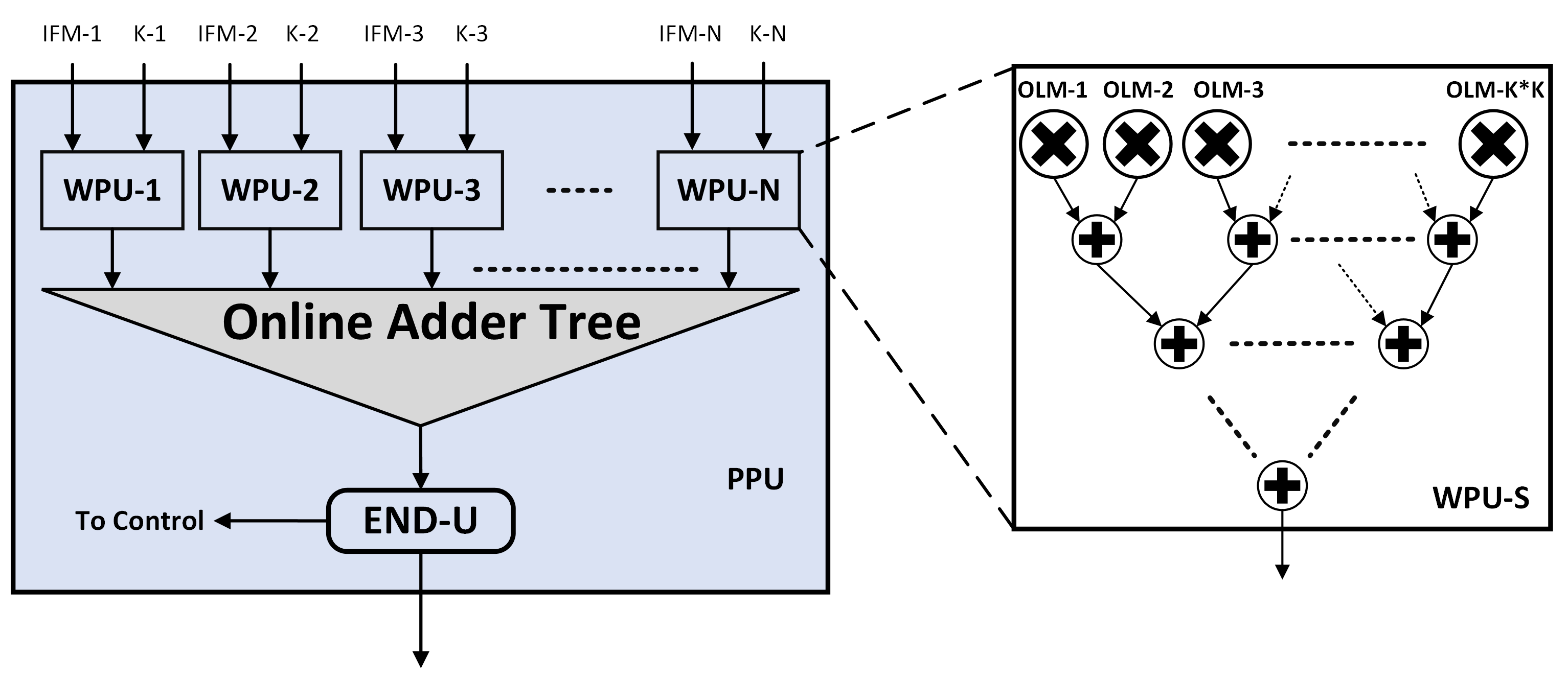}
    \caption{Internal Architecture of the Proposed Pixel Processing Unit with the window processing unit (WPU-S) that performs convolution in a spatial manner}
    \label{fig:PPU}
\end{figure}

Each PPU supports input tiling \cite{zhang2015optimizing} by the provision of ($t_n = N$) window processing units (WPU-S), where each WPU-S is responsible to generate the inner product of $K \times K$ pixels from one of the $N$ input features maps. The output of each WPU-S is forwarded to an adder tree which results in one output pixel in one of the output feature maps. Each PPU also contains an early negative detection unit (END-U) responsible for the detection and generating control signals if the output of a PPU is going to result in a negative value. The architecture of the PPU is presented in Fig.~\ref{fig:PPU}.

\subsubsection{Design Strategy-2 (DS-2) - Temporal Design} \label{sec:online-temporal}
\begin{figure}[!ht]
    \centering
    \includegraphics[width=0.45\linewidth]{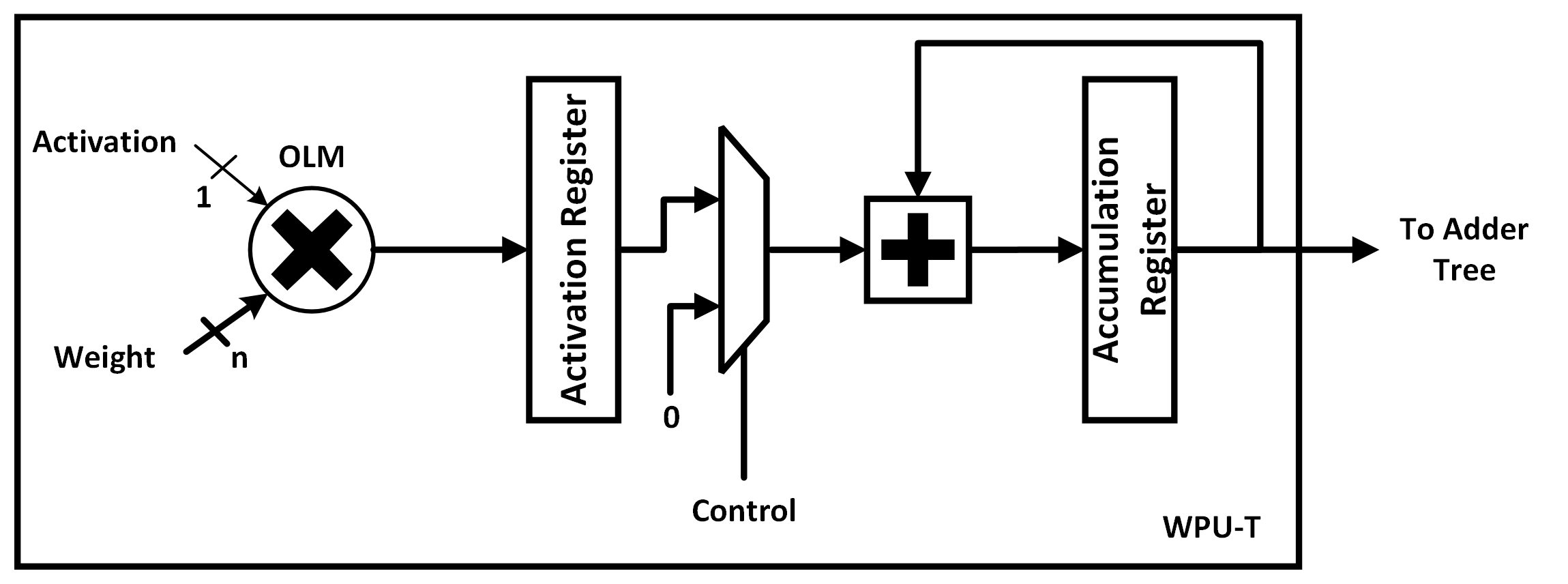}
    \caption{Architecture of the proposed window processing unit (WPU-T) that leverages the temporal computation pattern in convolution}
    \label{fig:WPU2}
\end{figure}

An alternate design is presented that aims to perform convolutions in a temporal fashion. Consequently the amount of basic computation units required to compute a $K\times K$ convolution window are reduced. In contrast to the WPU-S in the PPU design presented in Fig.~\ref{fig:PPU}, the window processing unit (WPU-T) in the present design allocates only one online arithmetic-based multiplier for the computation of a convolution window. This computation is carried out such that the online multiplier (OLM) is followed by an activation register that collects and stacks these output digits until all the output digits pertaining to one multiplication have been collected in the activation register. The contents of the activation register are then forwarded to an accumulation buffer until the results of the $k\times K$ multiplications have been accumulated. The contents of this accumulation register are then forwarded, in an MSDF manner, to an online arithmetic-based adder tree responsible to generate the sum across the $N$ input channels, ultimately resulting in the final output to be forwarded to the next operation in the CNN. It is also worth noting that the WPU-T pertaining to the temporal design can be replaced with WPU-S used in the PPU design presented in Fig.~\ref{fig:PPU}. The architecture of the WPU-T that leverages the temporal computation pattern is presented in Fig.~\ref{fig:WPU2}.

\section{Experimental Results and Discussion} \label{sec:4}
This section presents the experimental setup, performance evaluation parameters, results, comparisons, and discussion on the results obtained after the evaluation of the proposed designs.

\subsection{Experimental Setup}

\begin{figure}
    \centering
    \includegraphics[width=0.40\linewidth]{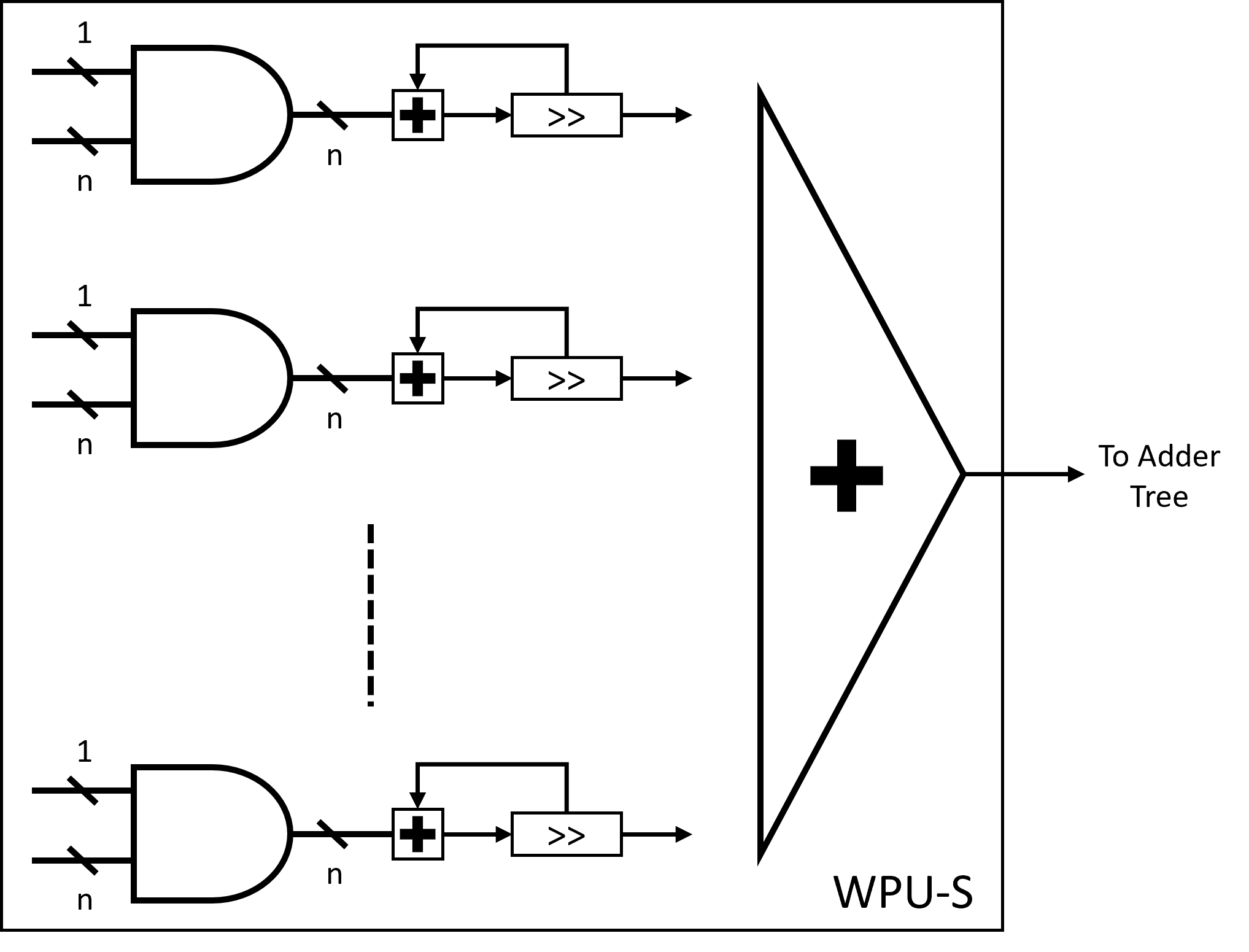}
    \caption{Architecture of the window processing unit (WPU-S), for conventional bit-serial design, that performs convolution of a $K\times K$ convolution window spatially.}
    \label{fig:bit-serial-wpus}
\end{figure}

In order to evaluate and compare the performance of the proposed designs with conventional bit-serial architectures, three baseline designs are used; 1) Baseline-1: conventional bit-serial design based on the processing element from UNPU \cite{lee2018unpu} with the \textit{tile stride} matching the convolution stride, 2) Baseline-2: online arithmetic-based design, also using the \textit{tile stride} as the convolution stride, and 3) Baseline-3: conventional bit-serial design where the \textit{tile stride} matches the proposed designs. All baselines utilze the same accelerator architecture and array layout as the proposed designs. The architecture for both baseline conventional bit-serial designs follows a similar structure to the proposed design. However, in conventional bit-serial designs, the window processing units (WPUs) use AND gate arrays for partial product generation, followed by an accumulator to sum the partial products. The WPU-S design for spatial design (DS-1) is shown in Fig.~\ref{fig:bit-serial-wpus}.

Each of the baseline designs use the same accelerator architecture and array layout as the proposed designs. The architecture of the conventional bit-serial arithmetic-based baseline designs for both the design strategies also follow a similar accelerator architecture as the proposed design. However, the design of the window processing units (WPUs) for both the design strategies of conventional bit-serial designs contain AND gate arrays for partial product generation, followed by an accumulator to obtain the sum of the partial products. The WPU-S design, for spatial design (DS-1), is presented in Fig.~\ref{fig:bit-serial-wpus}. The accumulation process in the figure handles the summing of partial products, while the subsequent adder tree computes the sum of $K \times K$ products. The resulting SoP from this adder tree is then passed to another adder tree, shown in the PPU in Fig.~\ref{fig:PPU}, which performs the final summation over $N$ input channels.


\begin{figure}[!ht]
    \centering
    \includegraphics[width=0.55\linewidth]{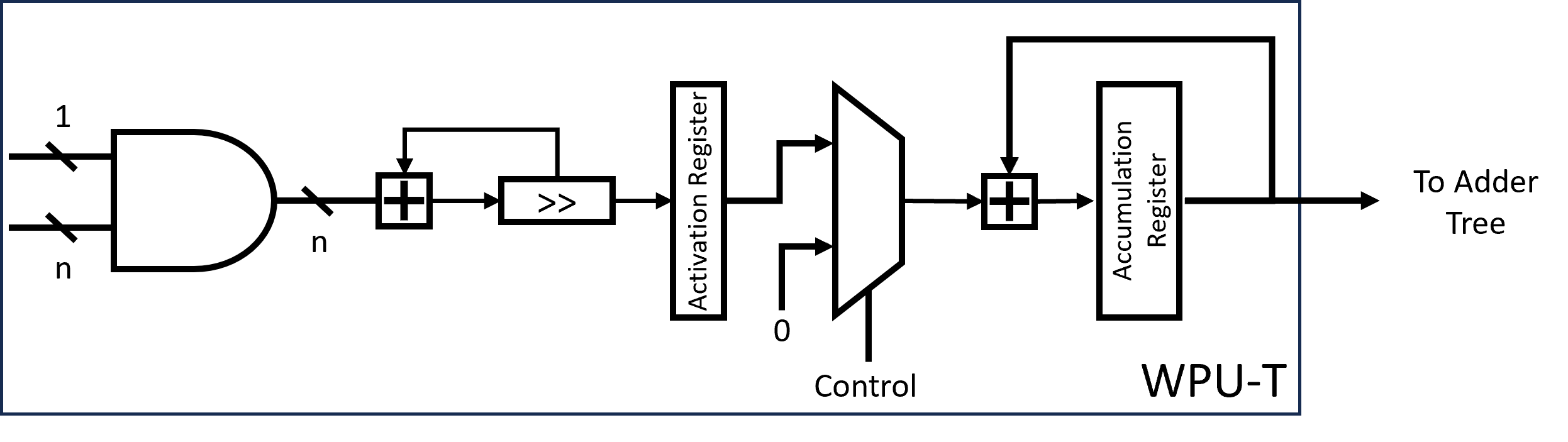}
    \caption{Architecture of the window processing unit (WPU-T), for conventional bit-serial design, that performs convolution of a $K\times K$ convolution window in a temporal fashion.}
    \label{fig:bit-serial-wput}
\end{figure}

In contrast to the spatial conventional bit-serial design presented in Fig.~\ref{fig:bit-serial-wpus}, a temporal design similar to that presented in Section \ref{sec:online-temporal} is also devised. The WPU-T architecture for conventional bit-serial design follows a similar strategy as presented in Fig.~\ref{fig:WPU2}, where the product of each of the $K\times K$ multiplications is carried out using a single multiplier. The architecture of the conventional bit-serial WPU-T that leverages the temporal computation pattern in convolution is shown in Fig.~\ref{fig:bit-serial-wput}.

In our experiments, we utilized three popular CNNs: LeNet-5 \cite{lecun1998gradient}, AlexNet \cite{krizhevsky2012imagenet}, and VGG-16 \cite{simonyan2014very}. For LeNet-5 and AlexNet, the first two convolution layers, along with their corresponding non-linear activation and pooling layers, were selected for fusion. In VGG-16, the first two convolution blocks, comprising four convolution layers, including their respective activation and pooling layers, were used for the fused-layer experiments.

 The RTL for the proposed and baseline accelerators was designed in Verilog and functionally verified using Xilinx Vivado 2023.2. We implemented the proposed designs on the Xilinx Virtex-7 VU19P FPGA. This FPGA platform was selected based on the availability of logic resources, as both the proposed and conventional bit-serial designs do not utilize built-in DSP resources for multiplication; instead, these resources are reserved for implementing the control units of the accelerators. \textcolor{black}{This is due to the fundamental architectural differences between conventional DSP operations and MSDF arithmetic. MSDF multipliers rely on a residual recurrence method rather than traditional partial product reduction. This approach requires cycle-to-cycle state tracking, and bidirectional digit propagation which are not supported by current DSP architectures. Consequently, we have developed the  MSDF-based arithmetic operators implemented using FPGA fabric resources.}
\subsection{Performance Evaluation Parameters}
The performance of the proposed method can be evaluated using various parameters such as performance, number of cycles, area, latency per image, inference speed-up, power efficiency, etc. The performance can be calculated using the following relation.
\begin{equation} \label{eq:perf}
    Performance = \frac{Num_{operations}}{number\ of\ execution\ cycles}
\end{equation}
Where, the number of operations ($Num_{operations}$) for a given convolution layer can be calculated as $2\times M\times N\times R\times C\times K\times K$. Where $M$ and $N$ represent the number of output and input feature maps respectively, $R$ and $C$ represent the height and width of the output feature map, and $K\times K$ is the dimension of the convolution kernel. Furthermore, the number of execution cycles (referred as $Cycles$ from here-on) in Eq.~\eqref{eq:perf} for the proposed online arithmetic-based design DS-1 can be calculated as;

\begin{equation} \label{eq:cyc1}
    \begin{aligned}
        Cycles = {} & \alpha^2 \times (\delta_{OLM} + \delta_{OLA}\times \ceil{\log{(K_1 \times K_1)}} 
        \\ & + \delta_{OLA}\times \ceil{\log{N_1}} + \ceil{\log{(K_1 \times K_1)}} + \ceil{\log{N_1}} 
        \\ & + MP_1 + \dots + \delta_{OLM} + \delta_{OLA}\times                         \ceil{\log{(K_Q \times K_Q)}} 
        \\ & + \delta_{OLA}\times \ceil{\log{N_Q}} +  \ceil{\log{(K_Q \times K_Q)}} 
        \\ & + \ceil{\log{N_Q}}  + MP_Q + n)  
    \end{aligned} 
\end{equation}

Where $\delta_{OLM}$ and $\delta_{OLA}$ represents the online delay for the multiplier and the adder respectively. These delays designate the number of cycles, usually up to $4$, that an online arithmetic-based component takes prior to generating the first digit (MSD) as its output. The expression $\ceil{\log{(K_Q \times K_Q)}}$ and $\ceil{\log{N_Q}}$ define the number of stages of the adder trees dedicated for computing the SoP for convolution window and input channels respectively for a convolution layer. $Q$ denotes the number of layers in the fusion design, $MP$ denotes the number of cycles required to perform the maxpooling operation, and $n$ denotes the precision of the input. Similarly, for the design DS-2, the number of cycles can be calculated as follows.

\begin{equation} \label{eq:cyc2}
    \begin{aligned}
        Cycles = {} & \alpha^2 \times ((\delta_{OLM} + (n-1) + Acc) \times K\times K 
        \\ & + \delta_{OLA}\times \ceil{\log{N_1}} + \ceil{\log{N_1}} + MP_1 
        \\ & + \dots + (\delta_{OLM} + (n-1) + Acc) \times K\times K 
        \\ & + \delta_{OLA}\times \ceil{\log{N_Q}} +  \ceil{\log{N_Q}}  + MP_Q + n)  
    \end{aligned} 
\end{equation}

Here $Acc$ denotes the number of cycles that the accumulator takes to perform the sum of $2$ operands. Both the relations also include the number of cycles elapsed due to the growth in the output precision due to the adder trees, and it is denoted by $\ceil{\log{(K \times K)}}$ and $\ceil{\log{N}}$ in the equations.

Other performance evaluation parameters are platform-specific such as, logic utilization, memory utilization, throughput, inference time per image, etc. The selection of implementation platform relies on the capacity of the hardware resources in coordination with the resource requirements of the accelerator design.
\subsection{Experimental Results}

\begin{figure}[!ht]
    \centering
    \includegraphics[width=0.65\linewidth]{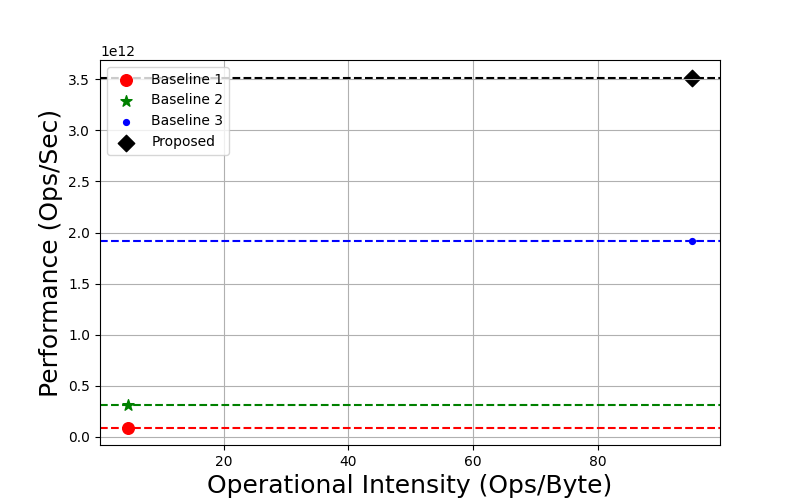}
    \caption{Performance vs. operational intensity comparison of the proposed spatial design (DS-1) with the baseline designs for the first convolution layer of AlexNet.}
    \label{fig:roofline_AN_L1}
\end{figure}

The proposed tile stride strategy coupled with the online arithmetic-based accelerator design not only improves the performance but can also improve the memory communication categorized by the operational intensity metric \cite{ofenbeck2014applying}. An analysis depicting the efficacy of the proposed technique is presented in Fig.~\ref{fig:roofline_AN_L1}. The figure shows that the proposed design and Baseline-3 design, using the proposed tile stride technique, have the same operational intensity as the other baseline designs. However, it is noteworthy that the performance of the proposed design surpasses that of Baseline-3 design. This demonstrates that the proposed tiling strategy, in combination with the superior capabilities of the online arithmetic paradigm, can outperform the conventional bit-serial design in terms of performance.

\begin{figure}[!ht]
\begin{center}
\begin{subfigure}{.65\linewidth}
  \centering
  \includegraphics[width=0.65\linewidth]{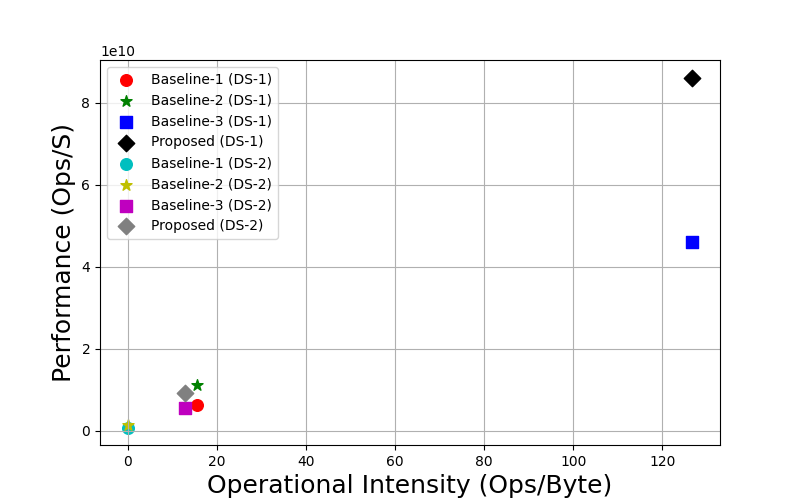}  
  \caption{LeNet-5}
  \label{fig:LN_fused}
\end{subfigure}
\begin{subfigure}{.65\linewidth}
  \centering
  \includegraphics[width=0.65\linewidth]{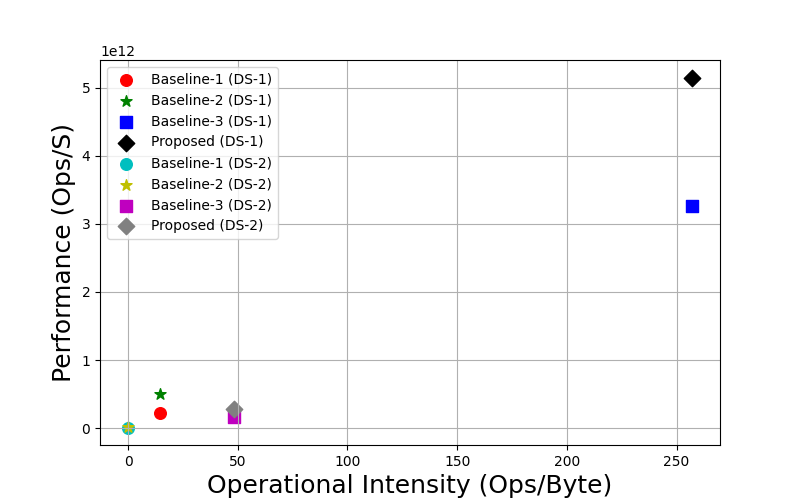}  
  \caption{AlexNet}
  \label{fig:AN_fused}
\end{subfigure}
\begin{subfigure}{.65\linewidth}
  \centering
  \includegraphics[width=0.65\linewidth]{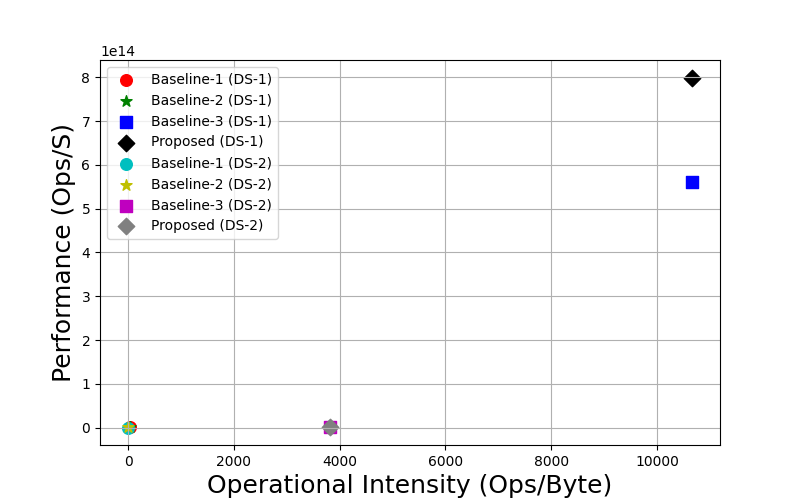}  
  \caption{VGG}
  \label{fig:VGG_fused}
\end{subfigure}
\end{center}
\caption{Performance vs. operational intensity comparison of the proposed spatial (DS-1) and temporal (DS-2) designs with the baseline designs for LeNet-5, AlexNet, and VGG models.}
\label{fig:roofline_fused}
\end{figure}

\renewcommand{\arraystretch}{1.2}
\begin{table*}[!ht] 
\centering
\caption{Performance comparison of the proposed Spatial design (DS-1) with the baseline designs. The duration and performance is listed in terms of micro seconds ($\mu s$) and TOPS unless specified otherwise.} \label{tab:perf-ds1}
\resizebox{\textwidth}{!}{
\begin{tabular}{|c|c|c|cc|cc|cc|cc|}
\hline
\multirow{2}{*}{\textbf{Network}} & \multirow{2}{*}{\textbf{Layer}} & \multirow{2}{*}{\textbf{Number of Operations}} & \multicolumn{2}{c|}{\textbf{Baseline-1}}                      & \multicolumn{2}{c|}{\textbf{Baseline-2}}                      & \multicolumn{2}{c|}{\textbf{Baseline-3}}                      & \multicolumn{2}{c|}{\textbf{Proposed}}                        \\ \cline{4-11} 
                                  &                                 &                                                & \multicolumn{1}{c|}{\textbf{Duration}} & \textbf{Performance} & \multicolumn{1}{c|}{\textbf{Duration}} & \textbf{Performance} & \multicolumn{1}{c|}{\textbf{Duration}} & \textbf{Performance} & \multicolumn{1}{c|}{\textbf{Duration}} & \textbf{Performance} \\ 
                                  \hline
\multirow{3}{*}{LeNet}            & CONV1                           & 235200                                         & \multicolumn{1}{c|}{138.72}            & 1.69 GOPS                 & \multicolumn{1}{c|}{57.80}              & 4.07 GOPS                 & \multicolumn{1}{c|}{12}                & 19.60 GOPS                 & \multicolumn{1}{c|}{5}                 & 47.04 GOPS                \\ \cline{2-11} 
                                  & CONV2                           & 940800                                         & \multicolumn{1}{c|}{41.31}             & 22.77 GOPS                & \multicolumn{1}{c|}{21.06}             & 44.67 GOPS                & \multicolumn{1}{c|}{12.75}             & 73.8 GOPS                & \multicolumn{1}{c|}{6.50}               & 144.74 GOPS               \\ \cline{2-11} 
                                  & Fused                           & 1183880                                        & \multicolumn{1}{c|}{187.43}            & 6.32 GOPS                & \multicolumn{1}{c|}{107.19}            & 11.04 GOPS                & \multicolumn{1}{c|}{25.75}             & 45.97 GOPS                & \multicolumn{1}{c|}{13.75}             & 86.10 GOPS                \\ \hline \hline
\multirow{3}{*}{AlexNet}          & CONV1                           & 105415200                                      & \multicolumn{1}{c|}{1109}              & 0.095                & \multicolumn{1}{c|}{623}               & 0.169                & \multicolumn{1}{c|}{53.46}             & 1.97                 & \multicolumn{1}{c|}{29.97}             & 3.517                \\ \cline{2-11} 
                                  & CONV2                           & 223948800                                      & \multicolumn{1}{c|}{337.50}             & 0.664                & \multicolumn{1}{c|}{268.75}            & 0.833                & \multicolumn{1}{c|}{43.74}             & 5.12                 & \multicolumn{1}{c|}{34.83}             & 6.43                 \\ \cline{2-11} 
                                  & Fused                           & 329659136                                      & \multicolumn{1}{c|}{1499.30}            & 0.219                & \multicolumn{1}{c|}{648.7}             & 0.508                & \multicolumn{1}{c|}{101.25}            & 3.26                 & \multicolumn{1}{c|}{63.99}             & 5.15                 \\ \hline \hline
\multirow{5}{*}{VGG}              & CONV1                           & 173408256                                      & \multicolumn{1}{c|}{8.11ms}            & 21.30 GOPS                & \multicolumn{1}{c|}{5.41ms}            & 32.10 GOPS                & \multicolumn{1}{c|}{3.78}              & 45.87                & \multicolumn{1}{c|}{2.52}              & 68.80                \\ \cline{2-11} 
                                  & CONV2                           & 3699376128                                     & \multicolumn{1}{c|}{9.14mS}            & 404.50 GOPS               & \multicolumn{1}{c|}{7.95mS}            & 465.30 GOPS               & \multicolumn{1}{c|}{4.14}              & 893.6                & \multicolumn{1}{c|}{3.60}               & 1027.60               \\ \cline{2-11} 
                                  & CONV3                           & 1849688064                                     & \multicolumn{1}{c|}{2.45mS}            & 754.56 GOPS              & \multicolumn{1}{c|}{2.13mS}            & 867.75 GOPS              & \multicolumn{1}{c|}{4.14}              & 446.8                & \multicolumn{1}{c|}{3.60}               & 513.80                \\ \cline{2-11} 
                                  & CONV4                           & 3699376128                                     & \multicolumn{1}{c|}{2.64mS}            & 1399.3 GOPS              & \multicolumn{1}{c|}{2.42mS}            & 1529.50 GOPS              & \multicolumn{1}{c|}{4.23}              & 874.56               & \multicolumn{1}{c|}{3.87}              & 955.90                \\ \cline{2-11} 
                                  & Fused                           & 9429625856                                     & \multicolumn{1}{c|}{23.36mS}           & 403.66 GOPS              & \multicolumn{1}{c|}{18.92mS}           & 498.40 GOPS               & \multicolumn{1}{c|}{16.83}             & 560.30                & \multicolumn{1}{c|}{11.79}             & 799.80                \\ \hline
\end{tabular}}
\end{table*}

Similarly, a comparison of performance vs. operational intensity of the fused-layer designs for LeNet-5, AlexNet, and VGG CNN models has been presented in Fig.~\ref{fig:roofline_fused}. The performance vs. operational intensity plots also confirm the findings presented in Fig.~\ref{fig:roofline_AN_L1}, that the proposed tile stride evaluation technique improves the operational intensity. For instance, the proposed spatial design (DS-1) improves the operational intensity for the LeNet-5, AlexNet, and VGG models by $8.20 \times$, $17.80 \times$, and $279.40 \times$, respectively. Similarly, utilizing online modules for arithmetic-based computations can result in significant performance enhancements, as demonstrated in Figs.~\ref{fig:LN_fused}, \ref{fig:AN_fused}, and \ref{fig:VGG_fused}.


As outlined in the experimental setup, we evaluate the proposed designs on LeNet-5, AlexNet, and VGG-16 networks. The performance and evaluation duration using the proposed design (DS-1) compared to the baseline designs are presented in Table~\ref{tab:perf-ds1}. All designs are evaluated at a frequency of $100$ MHz, with inference time (referred to as duration) and performance listed in the table. Notably, online arithmetic-based designs consistently outperform conventional bit-serial designs, regardless of the tile stride strategy. Specifically, the fused layer design based on online arithmetic achieves performance improvements of $1.75\times$, $2.32\times$, and $1.23\times$ for LeNet-5, AlexNet, and VGG, respectively, without the proposed tile stride strategy. When using the proposed tile stride technique, the online arithmetic design outperforms Baseline-3 by achieving $1.87\times$, $1.58\times$, and $1.43\times$ superior performance for LeNet-5, AlexNet, and VGG, respectively.

\renewcommand{\arraystretch}{1.2}
\begin{table}[!ht]
\centering
\caption{Performance comparison of the proposed Temporal design (DS-2) with the conventional bit-serial design (Baseline-3) using the proposed tile stride technique. The duration and performance is listed in terms of milli seconds ($ms$) and GOPS unless specified otherwise.} \label{tab:perf-ds2}
\resizebox{\linewidth}{!}{
\begin{tabular}{|c|c|c|cc|cc|}
\hline
\multirow{2}{*}{\textbf{Network}} & \multirow{2}{*}{\textbf{Layer}} & \multirow{2}{*}{\textbf{Number of Operations}} & \multicolumn{2}{c|}{\textbf{Baseline-3}}                               & \multicolumn{2}{c|}{\textbf{Proposed}}                                 \\ \cline{4-7} 
                                  &                                 &                                                & \multicolumn{1}{c|}{\textbf{Duration}}          & \textbf{Performance} & \multicolumn{1}{c|}{\textbf{Duration}}          & \textbf{Performance} \\ \hline
\multirow{3}{*}{LeNet}            & CONV1                           & 235200                                         & \multicolumn{1}{c|}{0.11}                       & 2.21                 & \multicolumn{1}{c|}{62.5$\mu S$}   & 3.80                  \\ \cline{2-7} 
                                  & CONV2                           & 940800                                         & \multicolumn{1}{c|}{0.11}                       & 8.80                  & \multicolumn{1}{c|}{64 $\mu S$}    & 14.70                 \\ \cline{2-7} 
                                  & Fused                           & 1183880                                        & \multicolumn{1}{c|}{0.21}                       & 5.53                 & \multicolumn{1}{c|}{128.25$\mu S$} & 9.20                  \\ \hline \hline
\multirow{3}{*}{AlexNet}          & CONV1                           & 105415200                                      & \multicolumn{1}{c|}{1.67}                       & 63.2             & \multicolumn{1}{c|}{0.983}                      & 107.20                \\ \cline{2-7} 
                                  & CONV2                           & 223948800                                      & \multicolumn{1}{c|}{0.35}                       & 641.50                & \multicolumn{1}{c|}{0.22}                       & 1039.40               \\ \cline{2-7} 
                                  & Fused                           & 329659136                                      & \multicolumn{1}{c|}{2.02}                       & 163.20                & \multicolumn{1}{c|}{1.21}                       & 273.50                \\ \hline \hline
\multirow{5}{*}{VGG}              & CONV1                           & 173408256                                      & \multicolumn{1}{c|}{13.95 $\mu S$} & 1243.10               & \multicolumn{1}{c|}{8.64 $\mu S$}  & 2007.04              \\ \cline{2-7} 
                                  & CONV2                           & 3699376128                                     & \multicolumn{1}{c|}{14.31 $\mu S$} & 258.50 TOPS           & \multicolumn{1}{c|}{9.72 $\mu S$}  & 380.60 TOPS           \\ \cline{2-7} 
                                  & CONV3                           & 1849688064                                     & \multicolumn{1}{c|}{14.31 $\mu S$} & 128.30 TOPS           & \multicolumn{1}{c|}{9.72}                       & 190.30 TOPS           \\ \cline{2-7} 
                                  & CONV4                           & 3699376128                                     & \multicolumn{1}{c|}{15.03 $\mu S$} & 246.10 TOPS           & \multicolumn{1}{c|}{9.90}                        & 370.30 TOPS           \\ \cline{2-7} 
                                  & Fused                           & 9429625856                                     & \multicolumn{1}{c|}{57.50 $\mu S$}  & 163.90 TOPS           & \multicolumn{1}{c|}{39.40 $\mu S$}  & 239.20 TOPS           \\ \hline
\end{tabular}}
\end{table}

For the temporal design DS-2, we present the comparative results of inference time and performance in-terms of GOPS for the conventional bit-serial design (Baseline-3) and the proposed design that use the proposed tile stride technique. Table~\ref{tab:perf-ds2} clearly shows that the proposed online arithmetic-based temporal design achieves $1.66\times$, $1.68\times$, and $1.46\times$ superior performance, in-terms of operations per second, for the fused layer designs of LeNet-5, AlexNet, and VGG respectively. The results presented in Tables \ref{tab:perf-ds1} and \ref{tab:perf-ds2} not only showcases the ability of online arithmetic-based designs over the conventional bit-serial arithmetic-based designs, but also confirm the utility of the proposed layer-fusion technique.

A comparison of the FPGA implementations of the proposed designs with the conventional bit-serial design (Baseline-3) is presented in Table~\ref{tab:FPGA-DS1} for the LeNet-5, AlexNet, and VGG models, all evaluated at a frequency of $100$ MHz. The results indicate that the proposed method utilizes more logic resources and BRAM compared to the baseline designs. However, for larger networks like VGG, the BRAM requirement for the proposed design is significantly lower than that of the baseline design. This reduction is attributed to the arithmetic nature of the proposed design, where output digits in MSDF format can be directly forwarded to the next processing units, minimizing the need for large intermediate buffers. Additionally, the proposed design achieves speedups of $1.87\times$, $1.58\times$, and $1.43\times$ for the implementations of LeNet-5, AlexNet, and VGG, respectively.


\renewcommand{\arraystretch}{1.2}
\begin{table*}[]
\centering
\caption{Comparison of FPGA implementation of proposed spatial design (DS-1) with the conventional bit-serial design with the proposed tiling scheme (Baseline-3). The FPGA device used for this experiment is Xilinx Ultrascale+ Vertix-7 VU19P.} \label{tab:FPGA-DS1}
\resizebox{\linewidth}{!}{
\begin{tabular}{|l|cc|cc|cc|}
\hline
Design                  & \multicolumn{1}{c|}{Baseline-3} & Proposed  & \multicolumn{1}{c|}{Baseline-3} & Proposed & \multicolumn{1}{c|}{Baseline-3} & Proposed \\ \hline
CNN Model               & \multicolumn{2}{c|}{LeNet-5}                                    & \multicolumn{2}{c|}{AlexNet}                                    & \multicolumn{2}{c|}{VGG}                                        \\ \hline 
Logic Utilization       & \multicolumn{1}{c|}{18.40K (0.21$\%$)}      & 28.80K (0.322$\%$)                         & \multicolumn{1}{c|}{5619.30K (63 $\%$)}    & 8645K (96.70 $\%$)                        & \multicolumn{1}{c|}{7091K (79.30 $\%$)}    & 7555.50K (94.5 $\%$)                      \\ \hline
BRAM Utilization    & \multicolumn{1}{c|}{2 ($0.05\%$)}        & 3 ($0.06\%$)                         & \multicolumn{1}{c|}{62 ($2.90\%$)}      & 113 ($5.20\%$)                           & \multicolumn{1}{c|}{740 ($34.30\%$)}    & 211 ($9.80\%$)                         \\ \hline
Throughput (TOPS)       & \multicolumn{1}{c|}{45.97 GOPS} & 86.10 GOPS                     & \multicolumn{1}{c|}{3.26}       & 5.15                          & \multicolumn{1}{c|}{560.30}      & 799.80                        \\ \hline
Latency/Image ($\mu s$) & \multicolumn{1}{c|}{25.75}      & 13.75                         & \multicolumn{1}{c|}{101.25}     & 63.99                         & \multicolumn{1}{c|}{16.83}      & 11.79                         \\ \hline
Speedup & \multicolumn{1}{c|}{1}      & 1.87$\times$                         & \multicolumn{1}{c|}{1}     & 1.58$\times$                         & \multicolumn{1}{c|}{1}      & 1.43$\times$                         \\ \hline
\end{tabular}}
\end{table*}

Similarly, the comparison of the proposed temporal design (DS-2) with the conventional bit-serial baseline design (Baseline-3) is presented in Table~\ref{tab:FPGA-DS2}. A similar trend in the BRAM utilization can be observed where for the VGG model fusion design, the proposed method requires nearly $5.2\times$ less BRAMs compared to the baseline design. This is due to the inherent property of the proposed online arithmetic-based design where the intermediate output digits can be used directly for the computation of the subsequent layer or operations. The results also show that the proposed temporal design achieves speedup of $1.67\times$, $1.68\times$, and $1.46\times$ for the implementation of LeNet-5, AlexNet, and VGG respectively, compared to the conventional bit-serial baseline design.

\begin{table*}[]
\centering
\caption{Comparison of FPGA implementation of proposed temporal design (DS-2) with the conventional bit-serial design with the proposed tiling scheme (Baseline-3). The FPGA device used for this experiment is Xilinx Ultrascale+ Vertix-7 VU19P.} \label{tab:FPGA-DS2}
\resizebox{\linewidth}{!}{
\begin{tabular}{|l|cc|cc|cc|}
\hline
Design                  & \multicolumn{1}{c|}{Baseline-3} & Proposed  & \multicolumn{1}{c|}{Baseline-3} & Proposed & \multicolumn{1}{c|}{Baseline-3} & Proposed \\ \hline
CNN Model               & \multicolumn{2}{c|}{LeNet-5}                                    & \multicolumn{2}{c|}{AlexNet}                                    & \multicolumn{2}{c|}{VGG}                                        \\ \hline 
Logic Utilization       & \multicolumn{1}{c|}{4.50K ($0.05\%$)}      & 14.20K (0.16$\%$)                         & \multicolumn{1}{c|}{277K (3.10$\%$)}    & 874.20K (9.80$\%$) & \multicolumn{1}{c|}{1270K (14.20$\%$)}    & 4012.20K (44.90 $\%$)                      \\ \hline
BRAM Utilization    & \multicolumn{1}{c|}{2 ($0.05\%$)}        & 2 ($0.05\%$)                         & \multicolumn{1}{c|}{44 ($2.04\%$)}      & 75 ($3.5\%$) & \multicolumn{1}{c|}{701 ($32.5\%$)}    & 134 ($6.21\%$)                         \\ \hline
Throughput (GOPS) & \multicolumn{1}{c|}{5.53} & 9.20 & \multicolumn{1}{c|}{163.20} & 273.50 & \multicolumn{1}{c|}{164 TOPS}      & 239 TOPS                         \\ \hline
Latency/Image ($\mu s$) & \multicolumn{1}{c|}{214.25}      & 128.25                         & \multicolumn{1}{c|}{2020.14}     & 1205.30                         & \multicolumn{1}{c|}{57.51}      & 39.42                         \\ \hline
Speedup & \multicolumn{1}{c|}{1}      & 1.67$\times$                         & \multicolumn{1}{c|}{1}     & 1.68$\times$                         & \multicolumn{1}{c|}{1}      & 1.46$\times$                         \\ \hline
\end{tabular}}
\end{table*}

\begin{figure}[!ht]
\centering
\begin{subfigure}{.55\linewidth}
  \centering
  \includegraphics[width=0.55\linewidth]{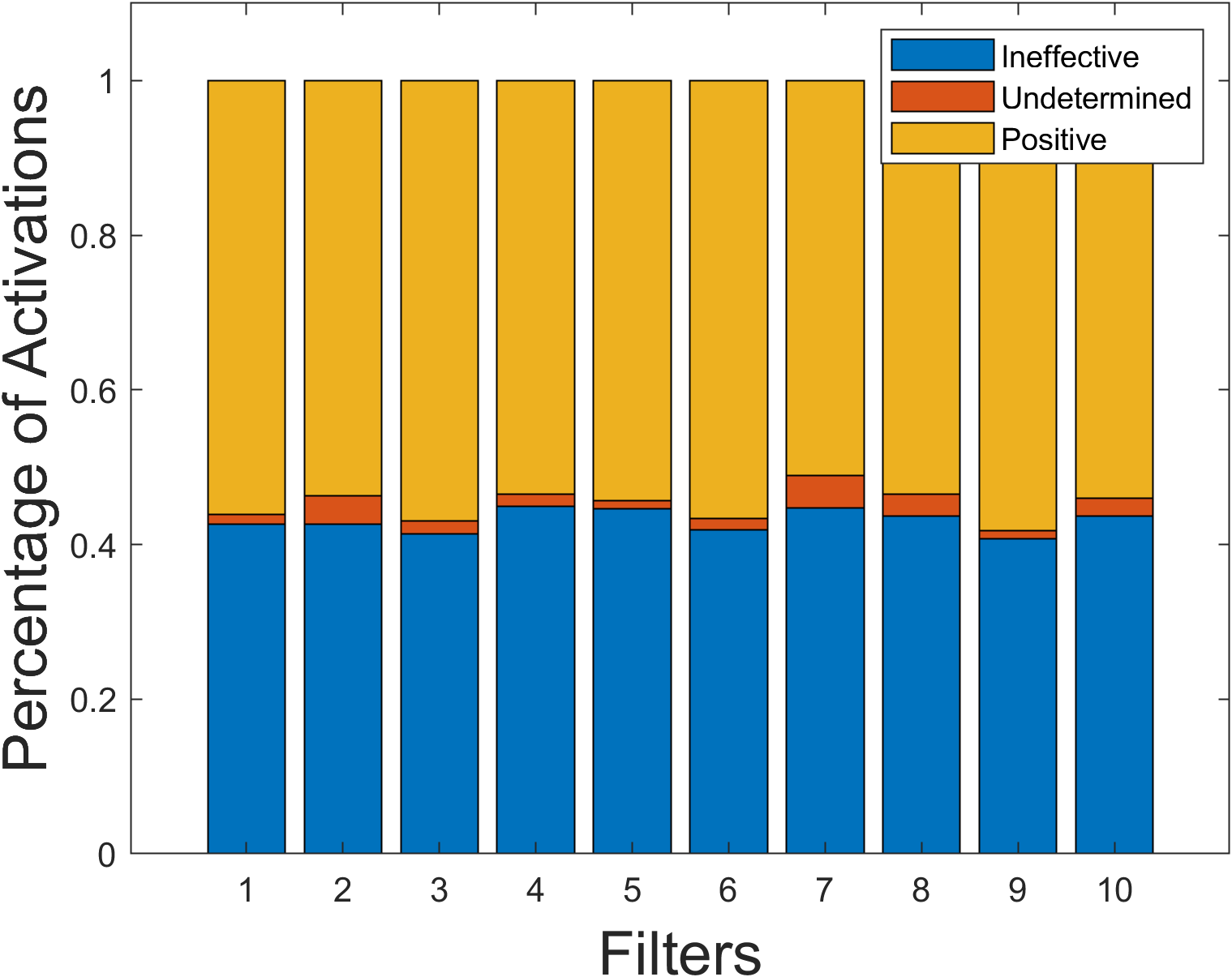}  
  \caption{AlexNet}
  \label{fig:AN_Neg}
\end{subfigure}
\begin{subfigure}{.55\linewidth}
  \centering
  \includegraphics[width=.55\linewidth]{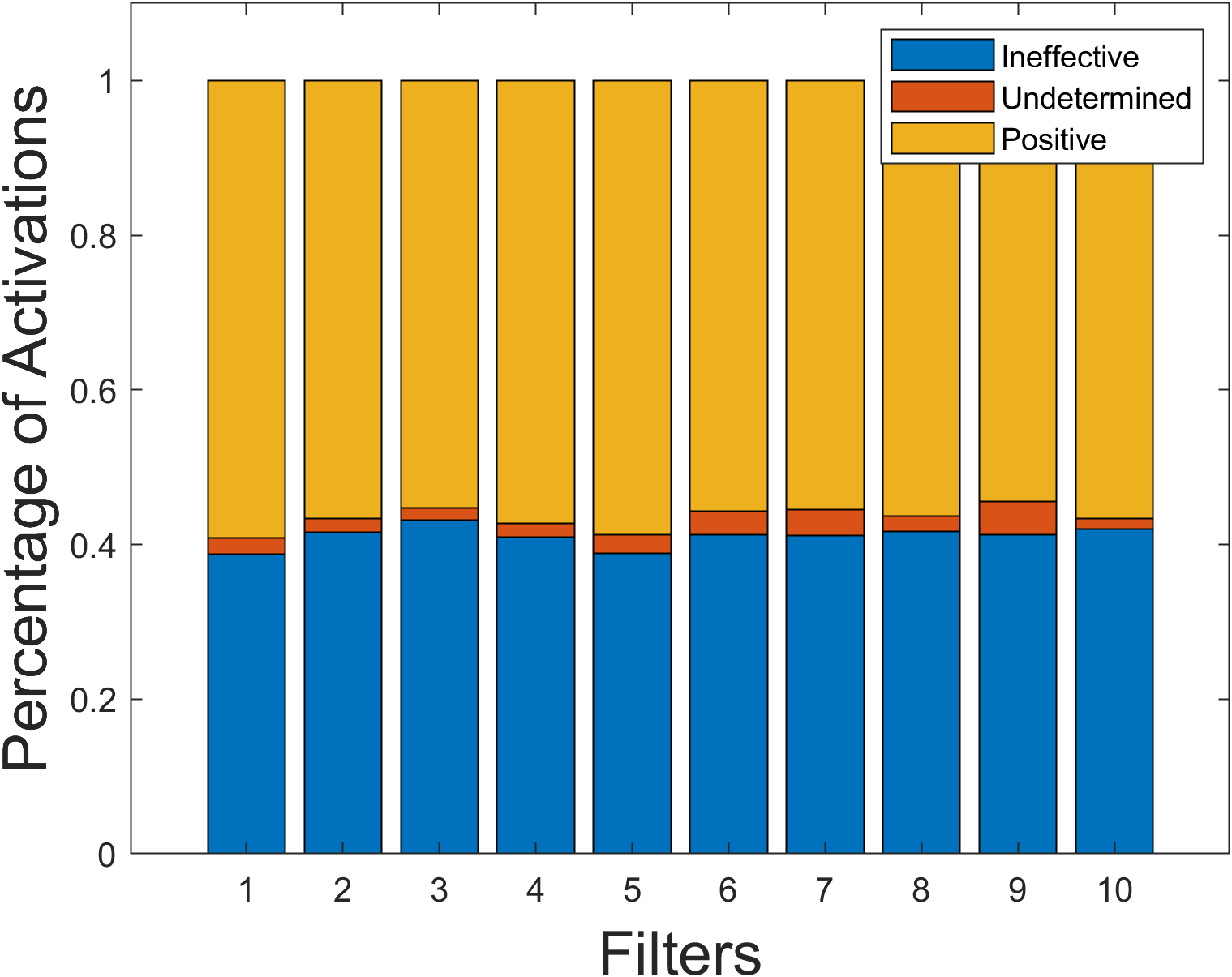}  
  \caption{VGG}
  \label{fig:VGG_Neg}
\end{subfigure}
\caption{Percentage of detected negative/ineffective activations for $10$ randomly selected filters in AlexNet and VGG models. The mean number of negative activations per output feature map is $43.10\%$ and $41.08\%$ for the first convolution layers of AlexNet and VGG respectively.}
\label{fig:early_neg}
\end{figure}

We also present the effect of early negative activation detection caused by ReLU activation function. For this experiment, we present the results of the proposed early negative detection technique on $10$ randomly selected filters for the first convolution layers of AlexNet and VGG models in Figs.~\ref{fig:AN_Neg} and \ref{fig:VGG_Neg} respectively. The analysis of the early negative detection technique show that an average of $43.1\%$ and $41.08\%$ activations per convolution filter were effectively determined as negative activations for the first convolution layers of AlexNet and VGG respectively. Nearly $2.36\%$ and $2.11\%$ activations were undetermined as either negative or positive. Upon examining the intermediate feature maps, it is determined that most of these undetermined activations were zero and hence did not cause any accuracy loss in the model classification performance. 

\begin{figure}[!ht]
\centering
\begin{subfigure}{.55\linewidth}
  \centering
  \includegraphics[width=.55\linewidth]{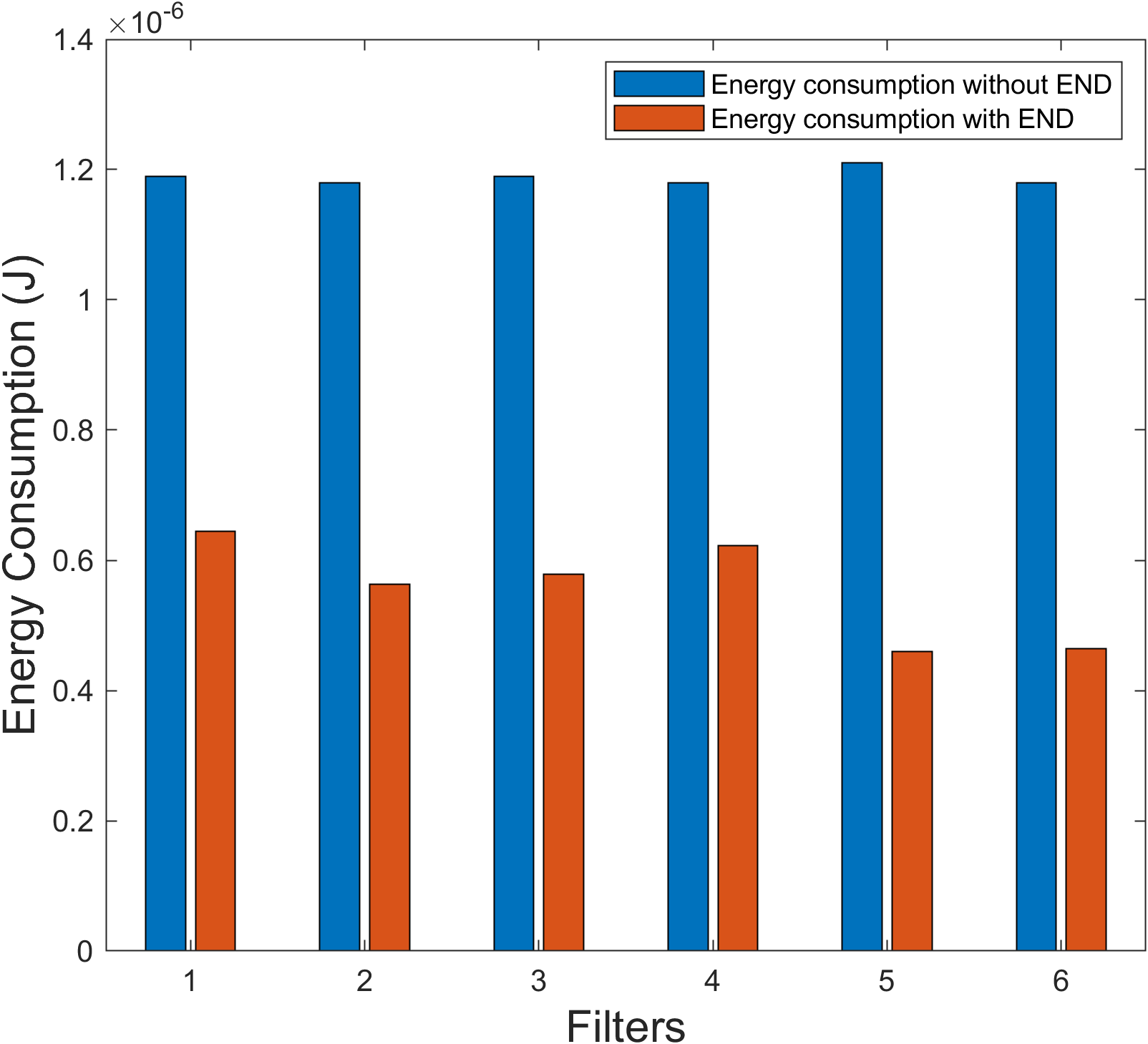}  
  \caption{LeNet-5}
  \label{fig:LN_energy}
\end{subfigure}
\begin{subfigure}{.55\linewidth}
  \centering
  \includegraphics[width=.55\linewidth]{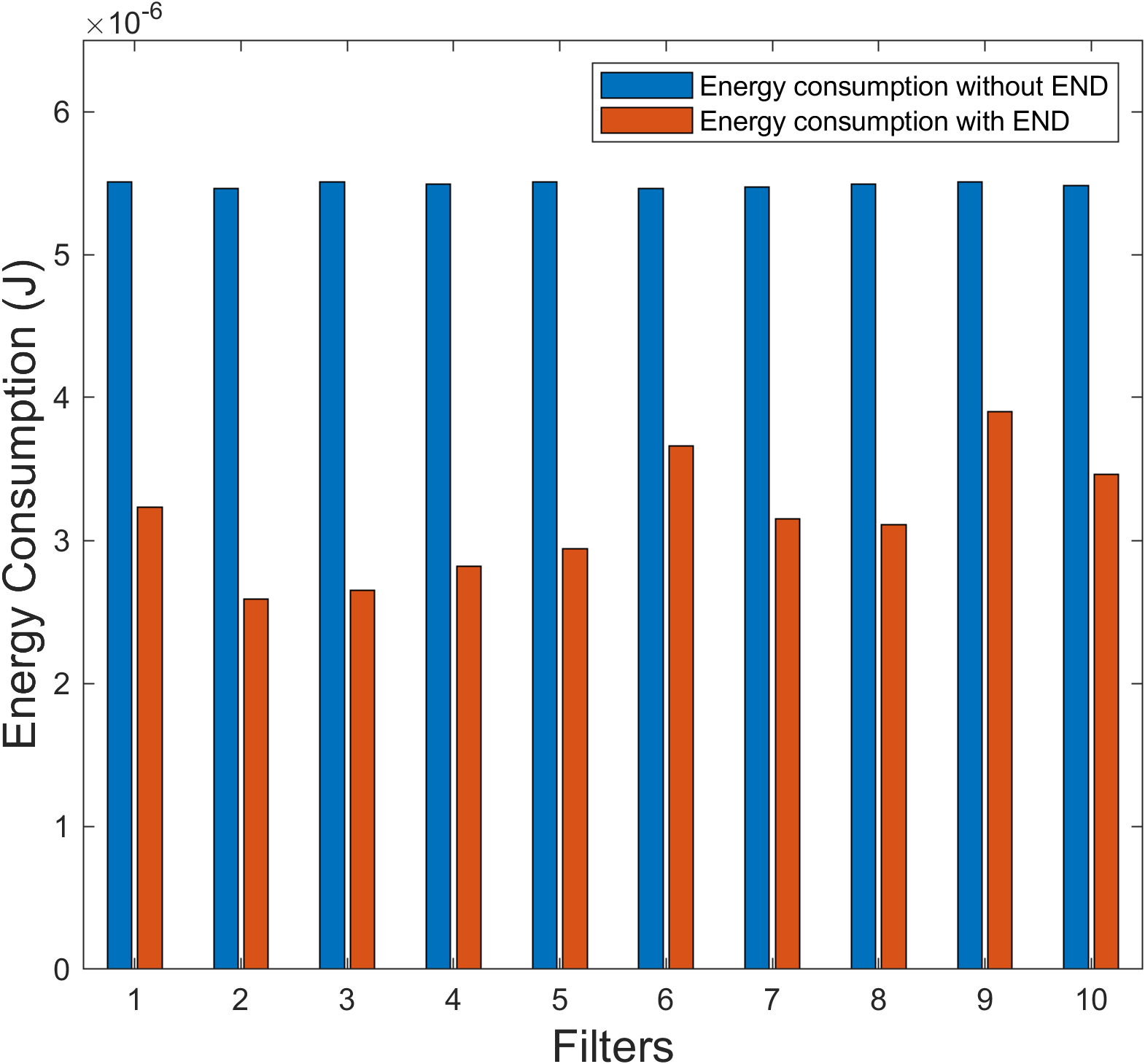}  
  \caption{AlexNet}
  \label{fig:AN_energy}
\end{subfigure}
\begin{subfigure}{.55\linewidth}
  \centering
  \includegraphics[width=.55\linewidth]{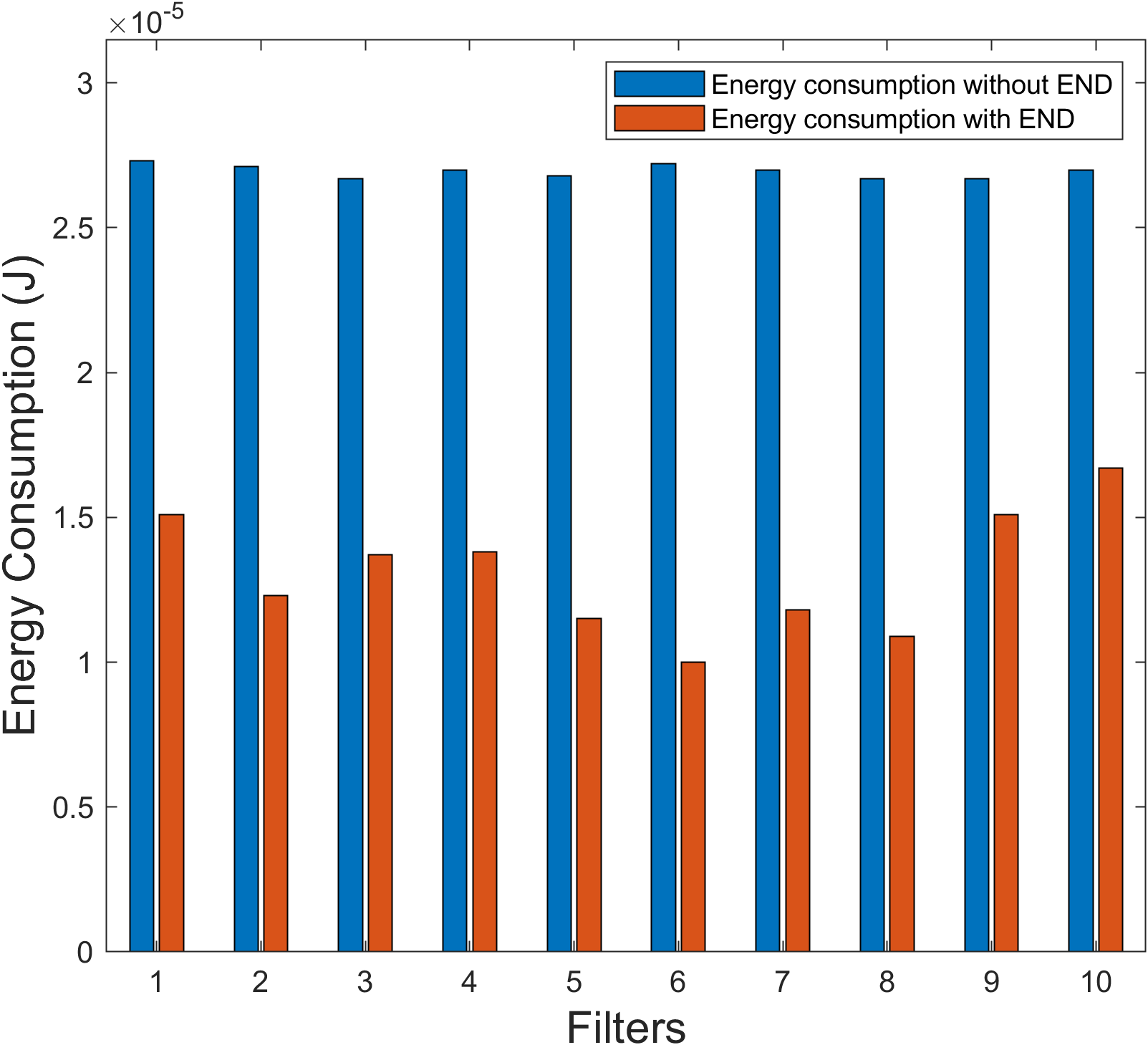}  
  \caption{VGG}
  \label{fig:VGG_energy}
\end{subfigure}
\caption{Energy savings with the proposed early negative detection (END) technique for the first convolution layers of LeNet-5, AlexNet, and VGG models. On average, $46.80\%$, $48.50\%$, and $42.60\%$ reduction in energy consumption is observed for LeNet-5, AlexNet, and VGG respectively.}
\label{fig:energy}
\end{figure}

Substantial energy savings can be achieved by detecting ineffective activations. In this context, results of the energy savings for the three networks used in this study are presented in Fig.~\ref{fig:energy}. The figure illustrates the energy consumption corresponding to $10$ randomly selected output feature maps of the first convolution layers. We performed our experiments with the proposed early negative detection (END) technique as well as without the proposed END technique using $10000$ images for all three networks. The proposed END technique resulted in substantial energy savings of $46.80\%$, $48.50\%$, and $42.60\%$ for LeNet-5, AlexNet, and VGG networks respectively. 

\begin{figure}[!ht]
    \centering
    \includegraphics[width = 0.45\linewidth]{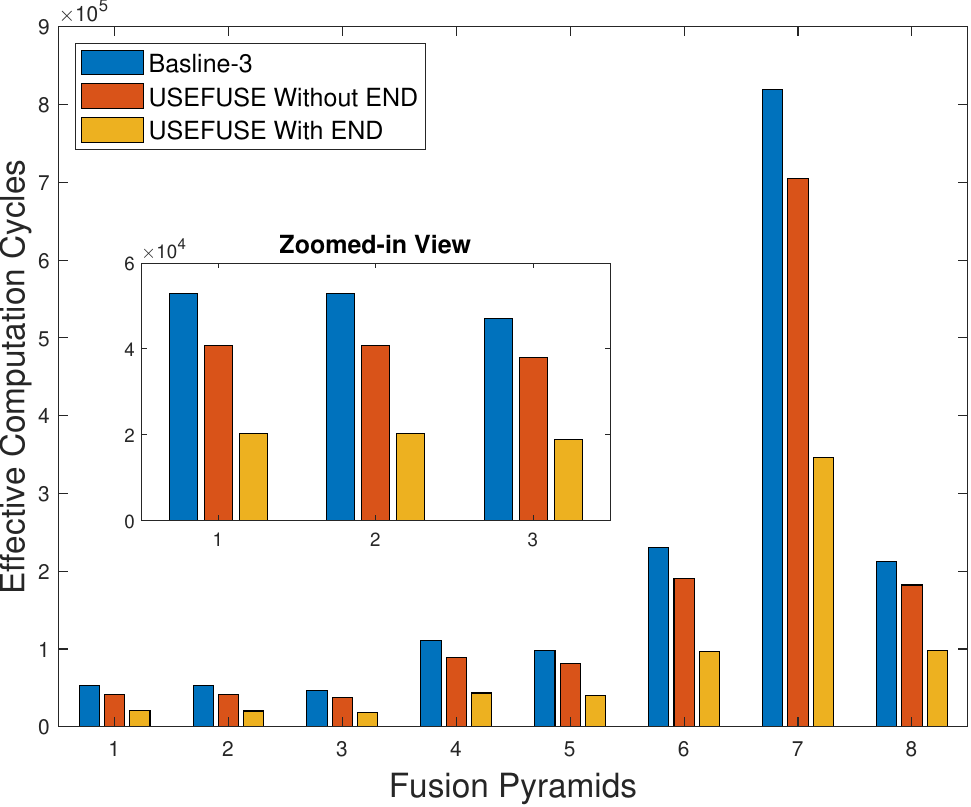}
    \caption{\textcolor{black}{The average effective computation cycles for each fusion pyramid were compared between the Baseline-3 design and the proposed design, with and without the END technique. The results showed that the END technique achieved an average savings of $50.1\%$ in computation cycles for the end-to-end flow.}}
    \label{fig:comp_cyc}
\end{figure}

Another experiment was conducted to demonstrate the effectiveness of the proposed END technique in reducing computation cycles within a fusion pyramid, using the ResNet-18 network. For this experiment, we fused two consecutive convolution layers, excluding the first convolution layer to ensure that each convolution block contains two fusion pyramids. We tested this setup on 100 images and report the average number of effective computation cycles with and without the proposed END scheme, for both the online arithmetic-based design and the conventional bit-serial (Baseline-3) design. The impact of the END technique on effective computation cycles is illustrated in Fig.~\ref{fig:comp_cyc}. It can be observed from the figure that the proposed END technique saves upto \(50.1\%\) cycles for the end-to-end execution of ResNet-18 workload using the proposed online arithmetic-based design. The comparison also shows the effectiveness of the online arithmetic-based computation where the online arithmetic designs with and without the END technique achieve $59.12\%$ and $18.4\%$ lower number of computation cycles compared to the conventional bit-serial design that uses the same accelerator architecture and the proposed tile stride technique. 

\subsection{Comparison with Previous Works}

{For the comparison with existing accelerators, we aim to accelerate the convolution layers in VGG-16 and ResNet-18 workloads in an end-to-end fashion. We conduct the performance comparison on several performance metrics such as latency per image, throughput in terms of GOPS, etc. For this experiment, we developed a fusion strategy by fusing 2 convolution layers in the fusion pyramid. The tiling parameters such as tile size, tile stride, etc., are calculated using Algorithms \ref{alg_tile_size} and \ref{alg_tile_stride}.}

\begin{table*}[!ht]
\centering
\caption{\textcolor{black}{Comparison with existing CNN accelerators. The baseline top-1 accuracy for VGG-16 and ResNet-18 are reported as $71.6\%$ and $69.76\%$ on their respective Pytorch \cite{paszke2019pytorch} websites.}} \label{tab:Comparison}
\resizebox{\linewidth}{!}{
\begin{tabular}{|l|cccc|c|cccc|c|}
\hline
\textbf{Model}                  & \multicolumn{5}{c|}{VGG-16}                                           & \multicolumn{5}{c|}{ResNet-18} \\ \hline
\textbf{Design}                 & \multicolumn{1}{c|}{TGPA \cite{wei2018tgpa}}  & \multicolumn{1}{c|}{\cite{ma2018automatic}} & \multicolumn{1}{c|}{\begin{tabular}[c]{@{}c@{}}ShortcutFusion\\ \cite{nguyen2022shortcutfusion} \end{tabular}} & \multicolumn{1}{c|}{\cite{hong2022accelerating}} & Proposed & \multicolumn{1}{c|}{\cite{latotzke2022design}} & \multicolumn{1}{c|}{T-DLA \cite{chen2019t}}   & \multicolumn{1}{c|}{\cite{xie2021efficient}} & \multicolumn{1}{c|}{RLDA \cite{fuketa2024multiplication}} & Proposed \\ \hline
\textbf{FPGA}                   & \multicolumn{1}{c|}{VU9P}  & \multicolumn{1}{c|}{\begin{tabular}[c]{@{}c@{}}Stratix 10\\ GX2800\end{tabular}} & \multicolumn{1}{c|}{KCU1500} & \multicolumn{1}{c|}{Alveo U50}  & VU5P & \multicolumn{1}{c|}{Stratix V}        & \multicolumn{1}{c|}{Zynq-7000}        & \multicolumn{1}{c|}{\begin{tabular}[c]{@{}c@{}}Arria10\\ SX660\end{tabular}} & \multicolumn{1}{c|}{\begin{tabular}[c]{@{}c@{}}Ultrascale+\\ XCZU7EV\end{tabular}} & VU5P \\ \hline
\textbf{Frequency (MHz)}        & \multicolumn{1}{c|}{210}   & \multicolumn{1}{c|}{300} & \multicolumn{1}{c|}{200} & \multicolumn{1}{c|}{200} & 100 & \multicolumn{1}{c|}{124}     & \multicolumn{1}{c|}{125}     & \multicolumn{1}{c|}{170} & \multicolumn{1}{c|}{150} & 100 \\ \hline
\textbf{Accuracy ($\%$)}        & \multicolumn{1}{c|}{-}   & \multicolumn{1}{c|}{-} & \multicolumn{1}{c|}{-}   & \multicolumn{1}{c|}{72.32}    & 71.21 & \multicolumn{1}{c|}{69.75}     & \multicolumn{1}{c|}{65.6}     & \multicolumn{1}{c|}{-} & \multicolumn{1}{c|}{65.5} & 69.13 \\ \hline
\textbf{Logic Utilization}      & \multicolumn{1}{c|}{493K}  & \multicolumn{1}{c|}{469K} & \multicolumn{1}{c|}{215.3K (33\%)} & \multicolumn{1}{c|}{601.7K (69\%)}         & 538.1K (89.5\%) & \multicolumn{1}{c|}{380.35K} & \multicolumn{1}{c|}{71.28\%} & \multicolumn{1}{c|}{102.6K} & \multicolumn{1}{c|}{230.4K (88.2\%)} & 542.6K (90.2\%) \\ \hline
\textbf{BRAM Utilization}       & \multicolumn{1}{c|}{3380}  & \multicolumn{1}{c|}{2421}  & \multicolumn{1}{c|}{1945 (45\%)}     & \multicolumn{1}{c|}{1084 (81\%)} & 1188 (58\%) & \multicolumn{1}{c|}{1644}    & \multicolumn{1}{c|}{68.93\%} & \multicolumn{1}{c|}{-} & \multicolumn{1}{c|}{307 (98.4\%)} & 1076 (52.54\%) \\ \hline
\textbf{Throughput (GOPS)}      & \multicolumn{1}{c|}{1510}  & \multicolumn{1}{c|}{1604.57} & \multicolumn{1}{c|}{607.5} & \multicolumn{1}{c|}{2895.5}    & 5594.7 & \multicolumn{1}{c|}{926.84}  & \multicolumn{1}{c|}{400}     & \multicolumn{1}{c|}{89.286} & \multicolumn{1}{c|}{620} & 1130.7 \\ \hline
\textbf{Latency per Image (ms)} & \multicolumn{1}{c|}{22.35} & \multicolumn{1}{c|}{19.29} & \multicolumn{1}{c|}{39.27} & \multicolumn{1}{c|}{13.90} & 9.18 & \multicolumn{1}{c|}{-}       & \multicolumn{1}{c|}{-}  & \multicolumn{1}{c|}{-} & \multicolumn{1}{c|}{-} & 14.44 \\ \hline
\end{tabular}}
\end{table*}

\textcolor{black}{The comparison of the proposed USEFUSE design with previous designs is presented in Table.~\ref{tab:Comparison}. The results presented in the table indicate that the proposed design uses comparatively large number of logic resources than its contemporary counterparts. However, the proposed USEFUSE design achieves $64.8\%$, $50.9\%$, and $38.9\%$ less BRAMs compared to TGPA \cite{wei2018tgpa}, \cite{ma2018automatic},and ShortcutFusion \cite{nguyen2022shortcutfusion} for VGG-16 workloads, respectively. For ResNet-18 workloads, USEFUSE uses $34.5\%$ less BRAM resources compared to the design presented in \cite{latotzke2022design}. The proposed design achieves significant throughput improvements of $3.7\times$, $3.48\times$, $9.2\times$, and $1.9\times$ for VGG-16 workloads compared to TGPA \cite{wei2018tgpa}, \cite{ma2018automatic}, ShortcutFusion \cite{nguyen2022shortcutfusion}, and \cite{hong2022accelerating} respectively. Similarly, USEFUSE achieved throughput improvements of $1.2\times$, $2.82\times$, $12.6\times$, and $1.82\times$ compared to the designs presented in \cite{latotzke2022design}, T-DLA \cite{chen2019t}, \cite{xie2021efficient}, and RDLA \cite{fuketa2024multiplication} respectively, for ResNet-18 workloads. Furthermore, the proposed design achieved $2.43\times$, $2.1\times$, $4.27\times$, and $1.5\times$ improvement in latency per image, compared to TGPA \cite{wei2018tgpa},  \cite{ma2018automatic}, ShortcutFusion \cite{nguyen2022shortcutfusion}, and \cite{hong2022accelerating} for VGG-16 workloads, respectively.}

The experimental results indicate that the use of online arithmetic-based compute units in the processing element can not only perform efficient computation of the convolution SOP, but also support the fusion of convolution layers in a CNN. Moreover, the MSDF nature of online arithmetic also aids in the early detection and subsequent termination of the ineffective computations that result in negative outputs. The proposed method of tile size and uniform stride calculation, coupled with online arithmetic-based compute units showcase superior performance compared to the state-of-the-art accelerator designs on VGG-16 and ResNet-18 workloads.

\section{Limitations and Future Work} \label{sec: Future}
While the proposed method offers significant advantages in terms of computational efficiency, it has certain limitations that we aim to address in future research.
Firstly, the proposed early negative detection technique limits the applicability to models relying on ReLU. While ReLU is fundamental and widely adopted, modern architectures also employ complex activation functions such as GELU, Sigmoid, Softmax, etc. Additionally, the proposed uniform stride method is specifically tested on ResNet-18, where skip connections are limited within individual residual blocks and do not span across multiple convolution blocks. This restriction allows for a simpler implementation of layer fusion, as the input from the skip connection can be integrated directly into the pipeline without requiring extensive reconfiguration. However, skip-connections spanning several convolution blocks may pose a challenge in determining the effective stride and tile size calculation which will be addressed in future research.

To overcome these limitations, our future work will focus on extending the method to support complex activation functions by developing efficient hardware implementations based on online arithmetic operators. This includes operations such as division, exponentiation, and power functions, which are commonly used in activation functions like Sigmoid and Softmax. This development will address both the challenges of accelerating modern architectures with complex activation functions and the need for efficient implementation of these same activation functions in the final output layers of neural network architectures to distinguish target applications. Therefore, the development of these activation functions will enhance the practicality of the proposed method by solving both issues simultaneously. 

Additionally, for architectures with longer skip connections, we propose integrating an adder within the pipeline to sum convolution outputs with skip connection inputs, requiring minimal structural changes and maintaining performance. Furthermore, a dynamic data flow control mechanism using multiplexers will be explored, allowing seamless switching between outputs from activation registers, skip connection registers, or zero values. These enhancements will allow the proposed accelerator to efficiently support a wider range of neural network models.

\paragraph{Extension to Modern Architectures}

\textcolor{black}{For transformer-based models, the computational dataflow significantly differs from CNNs, as multiple tokens are processed in parallel, and several attention heads operate simultaneously. While our current approach is primarily optimized for CNNs, its underlying principles can be extended to optimize transformer workloads. Specifically, the attention mechanism, which involves a sequence of dependent operations, could benefit from our MSDF mode of operation by enabling efficient pipelining. By restructuring the computation flow to exploit temporal parallelism, our approach could contribute to the acceleration of self-attention mechanisms. As part of our future research, we aim to explore tailored acceleration strategies for both depthwise convolutions in MobileNet and self-attention mechanisms in transformers, thereby extending the applicability of our method beyond CNNs.}

\paragraph{Hardware Optimizations for Low-Resource Deployment}
\textcolor{black}{To enhance the feasibility of our design for deployment on low-resource edge devices, several hardware optimizations can be explored. Our proposed temporal design (DS-2) reduces logic utilization by reusing computational resources over multiple cycles, and additional efficiency gains can be achieved through composite MSDF arithmetic operators. By designing a single SOP unit, we can minimize both logic area and latency, effectively decreasing the online delay while maintaining performance.}

\textcolor{black}{Furthermore, incorporating quantization and sparsity-aware optimizations can significantly reduce on-chip memory requirements. In our design, storage is already structured in multiples of 8-bit precision, making it inherently compatible with quantization techniques. Reducing precision lowers both memory footprint and compute latency without substantial accuracy degradation. Additionally, sparsity-aware optimizations can further decrease BRAM utilization by eliminating redundant computations and avoiding unnecessary storage of zero-valued parameters. Adaptive tiling strategies can be employed to maximize data reuse, thereby minimizing on-chip memory overhead for edge-device deployments. Moreover, resource-sharing mechanisms can optimize memory access patterns, ensuring efficient use of available storage.}
\textcolor{black}{These optimizations, combined with our proposed design principles, pave the way for high-performance yet resource-efficient deep learning accelerators, particularly suited for edge computing applications.}

By addressing these challenges, we aim to enhance the practicality and versatility of the proposed uniform stride and tiling method, enabling the accelerator to cater to a wide range of applications such as classification, detection, and segmentation.

\section{Conclusion} \label{sec: Conclusion}
This study introduces the use of low-latency left-to-right bit-serial arithmetic-based SOP units for convolution in fused CNN accelerators. Two designs cater to varied demands, emphasizing minimal response time (DS-1) for mission-critical applications and resource-constrained devices (DS-2). DS-1, a spatial computation pattern-based design, enhances operational intensity by $8.20 \times$, $17.80 \times$, and $279.40 \times$ for LeNet-5, AlexNet, and VGG networks, respectively. The temporal computation pattern-based design achieves speedups of $1.67 \times$, $1.68 \times$, and $1.46 \times$ for LeNet-5, AlexNet, and VGG networks respectively, surpassing conventional bit-serial baselines. An effective mechanism skips inefficient convolutions after ReLU layers, reducing power consumption without accuracy loss which  demonstrates substantial energy savings of $46.80\%$, $48.50\%$, and $42.60\%$ for LeNet-5, AlexNet, and VGG networks, respectively. Furthermore, the proposed USEFUSE has also exhibited superior performance compared to the existing CNN accelerator designs. These results underscore the efficacy of the proposed Uniform Stride strategy  for an improved operational intensity and optimizing energy consumption and computational speed in neural network implementations.

\bibliographystyle{unsrt}
\bibliography{references}

\begin{thebibliography}{10}

\bibitem{sun2020automatically}
Yanan Sun, Bing Xue, Mengjie Zhang, Gary~G Yen, and Jiancheng Lv.
\newblock Automatically designing cnn architectures using the genetic algorithm for image classification.
\newblock {\em IEEE transactions on cybernetics}, 50(9):3840--3854, 2020.

\bibitem{long2015fully}
Jonathan Long, Evan Shelhamer, and Trevor Darrell.
\newblock Fully convolutional networks for semantic segmentation.
\newblock In {\em Proceedings of the IEEE conference on computer vision and pattern recognition}, pages 3431--3440, 2015.

\bibitem{yoon2018efficient}
Yeo~Hun Yoon, Shujaat Khan, Jaeyoung Huh, and Jong~Chul Ye.
\newblock Efficient b-mode ultrasound image reconstruction from sub-sampled rf data using deep learning.
\newblock {\em IEEE transactions on medical imaging}, 38(2):325--336, 2018.

\bibitem{usman2022afp}
Muhammad Usman, Shujaat Khan, Seongyong Park, and Abdul Wahab.
\newblock Afp-src: identification of antifreeze proteins using sparse representation classifier.
\newblock {\em Neural Computing and Applications}, pages 1--11, 2022.

\bibitem{chen2022citisen}
Yu-Wen Chen, Kuo-Hsuan Hung, You-Jin Li, Alexander Chao-Fu Kang, Ya-Hsin Lai, Kai-Chun Liu, Szu-Wei Fu, Syu-Siang Wang, and Yu~Tsao.
\newblock Citisen: A deep learning-based speech signal-processing mobile application.
\newblock {\em IEEE Access}, 10:46082--46099, 2022.

\bibitem{chen2021dnnoff}
Xing Chen, Ming Li, Hao Zhong, Yun Ma, and Ching-Hsien Hsu.
\newblock Dnnoff: offloading dnn-based intelligent iot applications in mobile edge computing.
\newblock {\em IEEE transactions on industrial informatics}, 18(4):2820--2829, 2021.

\bibitem{lin2023intermittent}
Chih-Chia Lin, Chia-Yin Liu, Chih-Hsuan Yen, Tei-Wei Kuo, and Pi-Cheng Hsiu.
\newblock Intermittent-aware neural network pruning.
\newblock In {\em 2023 60th ACM/IEEE Design Automation Conference (DAC)}, pages 1--6. IEEE, 2023.

\bibitem{oh2022non}
Sangyun Oh, Hyeonuk Sim, Jounghyun Kim, and Jongeun Lee.
\newblock Non-uniform step size quantization for accurate post-training quantization.
\newblock In {\em European Conference on Computer Vision}, pages 658--673. Springer, 2022.

\bibitem{danopoulos2022adapt}
Dimitrios Danopoulos, Georgios Zervakis, Kostas Siozios, Dimitrios Soudris, and J{\"o}rg Henkel.
\newblock Adapt: Fast emulation of approximate dnn accelerators in pytorch.
\newblock {\em IEEE Transactions on Computer-Aided Design of Integrated Circuits and Systems}, 2022.

\bibitem{yoo20151}
Hoi-Jun Yoo, Seongwook Park, Kyeongryeol Bong, Dongjoo Shin, Jinmook Lee, and Sungpill Choi.
\newblock A 1.93 tops/w scalable deep learning/inference processor with tetra-parallel mimd architecture for big data applications.
\newblock In {\em IEEE international solid-state circuits conference}, pages 80--81. IEEE, 2015.

\bibitem{du2015shidiannao}
Zidong Du, Robert Fasthuber, Tianshi Chen, Paolo Ienne, Ling Li, Tao Luo, Xiaobing Feng, Yunji Chen, and Olivier Temam.
\newblock Shidiannao: Shifting vision processing closer to the sensor.
\newblock In {\em Proceedings of the 42nd Annual International Symposium on Computer Architecture}, pages 92--104, 2015.

\bibitem{chen2016eyeriss}
Yu-Hsin Chen, Tushar Krishna, Joel~S Emer, and Vivienne Sze.
\newblock Eyeriss: An energy-efficient reconfigurable accelerator for deep convolutional neural networks.
\newblock {\em IEEE journal of solid-state circuits}, 52(1):127--138, 2016.

\bibitem{judd2016stripes}
Patrick Judd, Jorge Albericio, Tayler Hetherington, Tor~M Aamodt, and Andreas Moshovos.
\newblock Stripes: Bit-serial deep neural network computing.
\newblock In {\em 2016 49th Annual IEEE/ACM International Symposium on Microarchitecture (MICRO)}, pages 1--12. IEEE, 2016.

\bibitem{lee2018unpu}
Jinmook Lee, Changhyeon Kim, Sanghoon Kang, Dongjoo Shin, Sangyeob Kim, and Hoi-Jun Yoo.
\newblock Unpu: An energy-efficient deep neural network accelerator with fully variable weight bit precision.
\newblock {\em IEEE Journal of Solid-State Circuits}, 54(1):173--185, 2018.

\bibitem{sharma2018bit}
Hardik Sharma, Jongse Park, Naveen Suda, Liangzhen Lai, Benson Chau, Vikas Chandra, and Hadi Esmaeilzadeh.
\newblock Bit fusion: Bit-level dynamically composable architecture for accelerating deep neural network.
\newblock In {\em 2018 ACM/IEEE 45th Annual International Symposium on Computer Architecture (ISCA)}, pages 764--775. IEEE, 2018.

\bibitem{kim2021compreend}
Namhyung Kim, Hanmin Park, Dongwoo Lee, Sungbum Kang, Jinho Lee, and Kiyoung Choi.
\newblock Compreend: Computation pruning through predictive early negative detection for relu in a deep neural network accelerator.
\newblock {\em IEEE Transactions on Computers}, 2021.

\bibitem{akhlaghi2018snapea}
Vahideh Akhlaghi, Amir Yazdanbakhsh, Kambiz Samadi, Rajesh~K Gupta, and Hadi Esmaeilzadeh.
\newblock Snapea: Predictive early activation for reducing computation in deep convolutional neural networks.
\newblock In {\em 2018 ACM/IEEE 45th Annual International Symposium on Computer Architecture (ISCA)}, pages 662--673. IEEE, 2018.

\bibitem{lee2018compend}
Dongwoo Lee, Sungbum Kang, and Kiyoung Choi.
\newblock Compend: Computation pruning through early negative detection for relu in a deep neural network accelerator.
\newblock In {\em Proceedings of the 2018 International Conference on Supercomputing}, pages 139--148, 2018.

\bibitem{chen2019comprrae}
Xizi Chen, Jingyang Zhu, Jingbo Jiang, and Chi-Ying Tsui.
\newblock Comprrae: Rram-based convolutional neural network accelerator with r educed computations through ar untime a ctivation e stimation.
\newblock In {\em Proceedings of the 24th Asia and South Pacific design automation conference}, pages 133--139, 2019.

\bibitem{ercegovac2004digital}
Milos~D Ercegovac and Tomas Lang.
\newblock {\em Digital arithmetic}.
\newblock Elsevier, 2004.

\bibitem{alwani2016fused}
Manoj Alwani, Han Chen, Michael Ferdman, and Peter Milder.
\newblock Fused-layer cnn accelerators.
\newblock In {\em 2016 49th Annual IEEE/ACM International Symposium on Microarchitecture (MICRO)}, pages 1--12. IEEE, 2016.

\bibitem{judd2018proteus}
Patrick Judd, Jorge Albericio, Tayler Hetherington, Tor Aamodt, Natalie~Enright Jerger, Raquel Urtasun, and Andreas Moshovos.
\newblock Proteus: Exploiting precision variability in deep neural networks.
\newblock {\em Parallel Computing}, 73:40--51, 2018.

\bibitem{shin2017fixed}
Sungho Shin, Yoonho Boo, and Wonyong Sung.
\newblock Fixed-point optimization of deep neural networks with adaptive step size retraining.
\newblock In {\em 2017 IEEE International conference on acoustics, speech and signal processing (ICASSP)}, pages 1203--1207. IEEE, 2017.

\bibitem{lu2021distilling}
Hang Lu, Liang Chang, Chenglong Li, Zixuan Zhu, Shengjian Lu, Yanhuan Liu, and Mingzhe Zhang.
\newblock Distilling bit-level sparsity parallelism for general purpose deep learning acceleration.
\newblock In {\em MICRO-54: 54th Annual IEEE/ACM International Symposium on Microarchitecture}, pages 963--976, 2021.

\bibitem{latotzke2022design}
Cecilia Latotzke, Tim Ciesielski, and Tobias Gemmeke.
\newblock Design of high-throughput mixed-precision cnn accelerators on fpga.
\newblock In {\em 2022 32nd International Conference on Field-Programmable Logic and Applications (FPL)}, pages 358--365. IEEE, 2022.

\bibitem{chen2019t}
Yao Chen, Kai Zhang, Cheng Gong, Cong Hao, Xiaofan Zhang, Tao Li, and Deming Chen.
\newblock T-dla: An open-source deep learning accelerator for ternarized dnn models on embedded fpga.
\newblock In {\em 2019 IEEE Computer Society Annual Symposium on VLSI (ISVLSI)}, pages 13--18. IEEE, 2019.

\bibitem{karadeniz2021talipot}
Mahmut~Burak Karadeniz and Mustafa Altun.
\newblock Talipot: Energy-efficient dnn booster employing hybrid bit parallel-serial processing in msb-first fashion.
\newblock {\em IEEE Transactions on Computer-Aided Design of Integrated Circuits and Systems}, 41(8):2714--2727, 2021.

\bibitem{liu2020precision}
Wenjian Liu, Jun Lin, and Zhongfeng Wang.
\newblock A precision-scalable energy-efficient convolutional neural network accelerator.
\newblock {\em IEEE Transactions on Circuits and Systems I: Regular Papers}, 67(10):3484--3497, 2020.

\bibitem{albericio2017bitpragmatic}
Jorge Albericio, Alberto Delmás, Patrick Judd, Sayeh Sharify, Gerard O’Leary, Roman Genov, and Andreas Moshovos.
\newblock Bit-pragmatic deep neural network computing.
\newblock In {\em 2017 50th Annual IEEE/ACM International Symposium on Microarchitecture (MICRO)}, pages 382--394, 2017.

\bibitem{al2020towards}
Khalid Al-Hawaj, Olalekan Afuye, Shady Agwa, Alyssa Apsel, and Christopher Batten.
\newblock Towards a reconfigurable bit-serial/bit-parallel vector accelerator using in-situ processing-in-sram.
\newblock In {\em 2020 IEEE International Symposium on Circuits and Systems (ISCAS)}, pages 1--5. IEEE, 2020.

\bibitem{waeijen2021convfusion}
Luc Waeijen, Savvas Sioutas, Maurice Peemen, Menno Lindwer, and Henk Corporaal.
\newblock Convfusion: A model for layer fusion in convolutional neural networks.
\newblock {\em IEEE Access}, 9:168245--168267, 2021.

\bibitem{zhao2018deepthings}
Zhuoran Zhao, Kamyar~Mirzazad Barijough, and Andreas Gerstlauer.
\newblock Deepthings: Distributed adaptive deep learning inference on resource-constrained iot edge clusters.
\newblock {\em IEEE Transactions on Computer-Aided Design of Integrated Circuits and Systems}, 37(11):2348--2359, 2018.

\bibitem{wei2018tgpa}
Xuechao Wei, Yun Liang, Xiuhong Li, Cody~Hao Yu, Peng Zhang, and Jason Cong.
\newblock Tgpa: Tile-grained pipeline architecture for low latency cnn inference.
\newblock In {\em 2018 IEEE/ACM International Conference on Computer-Aided Design (ICCAD)}, pages 1--8. ACM, 2018.

\bibitem{xiao2017exploring}
Qingcheng Xiao, Yun Liang, Liqiang Lu, Shengen Yan, and Yu-Wing Tai.
\newblock Exploring heterogeneous algorithms for accelerating deep convolutional neural networks on fpgas.
\newblock In {\em Proceedings of the 54th Annual Design Automation Conference 2017}, pages 1--6, 2017.

\bibitem{li2022fused}
Mingze Li, Ning Wang, Huan Zhou, Yubin Duan, and Jie Wu.
\newblock Fused-layer-based dnn model parallelism and partial computation offloading.
\newblock In {\em GLOBECOM 2022-2022 IEEE Global Communications Conference}, pages 5195--5200. IEEE, 2022.

\bibitem{zhou2022accelerating}
Huan Zhou, Mingze Li, Ning Wang, Geyong Min, and Jie Wu.
\newblock Accelerating deep learning inference via model parallelism and partial computation offloading.
\newblock {\em IEEE Transactions on Parallel and Distributed Systems}, 34(2):475--488, 2022.

\bibitem{cai2021olympus}
Xuyi Cai, Ying Wang, Kaijie Tu, Chengsi Gao, and Lei Zhang.
\newblock Olympus: Reaching memory-optimality on dnn processors.
\newblock {\em IEEE Transactions on Computers}, 71(8):1939--1951, 2021.

\bibitem{tewari2021minimizing}
Saurabh Tewari, Anshul Kumar, and Kolin Paul.
\newblock Minimizing off-chip memory access for cnn accelerators.
\newblock {\em IEEE Consumer Electronics Magazine}, 11(3):95--104, 2021.

\bibitem{ahmad2020superslash}
Hazoor Ahmad, Tabasher Arif, Muhammad~Abdullah Hanif, Rehan Hafiz, and Muhammad Shafique.
\newblock Superslash: A unified design space exploration and model compression methodology for design of deep learning accelerators with reduced off-chip memory access volume.
\newblock {\em IEEE Transactions on Computer-Aided Design of Integrated Circuits and Systems}, 39(11):4191--4204, 2020.

\bibitem{tewari2020bus}
Saurabh Tewari, Anshul Kumar, and Kolin Paul.
\newblock Bus width aware off-chip memory access minimization for cnn accelerators.
\newblock In {\em 2020 IEEE Computer Society Annual Symposium on VLSI (ISVLSI)}, pages 240--245. IEEE, 2020.

\bibitem{kang2021multi}
Duseok Kang, Donghyun Kang, and Soonhoi Ha.
\newblock Multi-bank on-chip memory management techniques for cnn accelerators.
\newblock {\em IEEE Transactions on Computers}, 71(5):1181--1193, 2021.

\bibitem{zhang2023practical}
Kai Zhang, Yawei Li, Jingyun Liang, Jiezhang Cao, Yulun Zhang, Hao Tang, Deng-Ping Fan, Radu Timofte, and Luc~Van Gool.
\newblock Practical blind image denoising via swin-conv-unet and data synthesis.
\newblock {\em Machine Intelligence Research}, 20(6):822--836, 2023.

\bibitem{li2024image}
Mingkang Li, Wei Liu, and Weidong Chen.
\newblock An image denoising method based on swin transformer v2 and u-net architecture.
\newblock In {\em 2024 IEEE 16th International Conference on Advanced Infocomm Technology (ICAIT)}, pages 204--209. IEEE, 2024.

\bibitem{zhang2021plug}
Kai Zhang, Yawei Li, Wangmeng Zuo, Lei Zhang, Luc Van~Gool, and Radu Timofte.
\newblock Plug-and-play image restoration with deep denoiser prior.
\newblock {\em IEEE Transactions on Pattern Analysis and Machine Intelligence}, 44(10):6360--6376, 2021.

\bibitem{chen2024artadapter}
Dar-Yen Chen, Hamish Tennent, and Ching-Wen Hsu.
\newblock Artadapter: Text-to-image style transfer using multi-level style encoder and explicit adaptation.
\newblock In {\em Proceedings of the IEEE/CVF Conference on Computer Vision and Pattern Recognition}, pages 8619--8628, 2024.

\bibitem{bhattad2024stylitgan}
Anand Bhattad, James Soole, and DA~Forsyth.
\newblock Stylitgan: Image-based relighting via latent control.
\newblock In {\em Proceedings of the IEEE/CVF Conference on Computer Vision and Pattern Recognition}, pages 4231--4240, 2024.

\bibitem{tafasca2024sharingan}
Samy Tafasca, Anshul Gupta, and Jean-Marc Odobez.
\newblock Sharingan: A transformer architecture for multi-person gaze following.
\newblock In {\em Proceedings of the IEEE/CVF Conference on Computer Vision and Pattern Recognition}, pages 2008--2017, 2024.

\bibitem{shen2023study}
Kai Shen, Junliang Guo, Xu~Tan, Siliang Tang, Rui Wang, and Jiang Bian.
\newblock A study on relu and softmax in transformer.
\newblock {\em arXiv preprint arXiv:2302.06461}, 2023.

\bibitem{wortsman2023replacing}
Mitchell Wortsman, Jaehoon Lee, Justin Gilmer, and Simon Kornblith.
\newblock Replacing softmax with relu in vision transformers.
\newblock {\em arXiv preprint arXiv:2309.08586}, 2023.

\bibitem{mallappa2022terminetor}
Uday Mallappa, Pranav Gangwar, Behnam Khaleghi, Haichao Yang, and Tajana Rosing.
\newblock Terminetor: Early convolution termination for efficient deep neural networks.
\newblock In {\em 2022 IEEE 40th International Conference on Computer Design (ICCD)}, pages 635--643. IEEE, 2022.

\bibitem{pan2023bitset}
Yunjie Pan, Jiecao Yu, Andrew Lukefahr, Reetuparna Das, and Scott Mahlke.
\newblock Bitset: Bit-serial early termination for computation reduction in convolutional neural networks.
\newblock {\em ACM Transactions on Embedded Computing Systems}, 22(5s):1--24, 2023.

\bibitem{asadikouhanjani2020novel}
Mohammadreza Asadikouhanjani and Seok-Bum Ko.
\newblock A novel architecture for early detection of negative output features in deep neural network accelerators.
\newblock {\em IEEE Transactions on Circuits and Systems II: Express Briefs}, 67(12):3332--3336, 2020.

\bibitem{shuvo2020msb}
Md~Kamruzzaman Shuvo, David~E Thompson, and Haibo Wang.
\newblock Msb-first distributed arithmetic circuit for convolution neural network computation.
\newblock In {\em 2020 IEEE 63rd International Midwest Symposium on Circuits and Systems (MWSCAS)}, pages 399--402. IEEE, 2020.

\bibitem{usman2023low}
Muhammad Usman, Milo{\v{s}} D.~Ercegovac, and Jeong-A Lee.
\newblock Low-latency online multiplier with reduced activities and minimized interconnect for inner product arrays.
\newblock {\em Journal of Signal Processing Systems}, pages 1--20, 2023.

\bibitem{zhang2015optimizing}
Chen Zhang, Peng Li, Guangyu Sun, Yijin Guan, Bingjun Xiao, and Jason Cong.
\newblock Optimizing fpga-based accelerator design for deep convolutional neural networks.
\newblock In {\em Proceedings of the 2015 ACM/SIGDA international symposium on field-programmable gate arrays}, pages 161--170, 2015.

\bibitem{lecun1998gradient}
Yann LeCun, L{\'e}on Bottou, Yoshua Bengio, and Patrick Haffner.
\newblock Gradient-based learning applied to document recognition.
\newblock {\em Proceedings of the IEEE}, 86(11):2278--2324, 1998.

\bibitem{krizhevsky2012imagenet}
Alex Krizhevsky, Ilya Sutskever, and Geoffrey~E Hinton.
\newblock Imagenet classification with deep convolutional neural networks.
\newblock {\em Advances in neural information processing systems}, 25, 2012.

\bibitem{simonyan2014very}
Karen Simonyan and Andrew Zisserman.
\newblock Very deep convolutional networks for large-scale image recognition.
\newblock {\em arXiv preprint arXiv:1409.1556}, 2014.

\bibitem{ofenbeck2014applying}
Georg Ofenbeck, Ruedi Steinmann, Victoria Caparros, Daniele~G Spampinato, and Markus P{\"u}schel.
\newblock Applying the roofline model.
\newblock In {\em 2014 IEEE International Symposium on Performance Analysis of Systems and Software (ISPASS)}, pages 76--85. IEEE, 2014.

\bibitem{paszke2019pytorch}
Adam Paszke, Sam Gross, Francisco Massa, Adam Lerer, James Bradbury, Gregory Chanan, Trevor Killeen, Zeming Lin, Natalia Gimelshein, Luca Antiga, et~al.
\newblock Pytorch: An imperative style, high-performance deep learning library.
\newblock {\em Advances in neural information processing systems}, 32, 2019.

\bibitem{ma2018automatic}
Yufei Ma, Yu~Cao, Sarma Vrudhula, and Jae-sun Seo.
\newblock Automatic compilation of diverse cnns onto high-performance fpga accelerators.
\newblock {\em IEEE Transactions on Computer-Aided Design of Integrated Circuits and Systems}, 39(2):424--437, 2018.

\bibitem{nguyen2022shortcutfusion}
Duy~Thanh Nguyen, Hyeonseung Je, Tuan~Nghia Nguyen, Soojung Ryu, Kyujoong Lee, and Hyuk-Jae Lee.
\newblock Shortcutfusion: From tensorflow to fpga-based accelerator with a reuse-aware memory allocation for shortcut data.
\newblock {\em IEEE Transactions on Circuits and Systems I: Regular Papers}, 69(6):2477--2489, 2022.

\bibitem{hong2022accelerating}
Seongmin Hong, Yashael~Faith Arthanto, Joo-Young Kim, et~al.
\newblock Accelerating deep convolutional neural networks using number theoretic transform.
\newblock {\em IEEE Transactions on Circuits and Systems I: Regular Papers}, 70(1):315--326, 2022.

\bibitem{xie2021efficient}
Xiaoru Xie, Jun Lin, Zhongfeng Wang, and Jinghe Wei.
\newblock An efficient and flexible accelerator design for sparse convolutional neural networks.
\newblock {\em IEEE Transactions on Circuits and Systems I: Regular Papers}, 68(7):2936--2949, 2021.

\bibitem{fuketa2024multiplication}
Hiroshi Fuketa, Toshihiro Katashita, Yohei Hori, and Masakazu Hioki.
\newblock Multiplication-free lookup-based cnn accelerator using residual vector quantization and its fpga implementation.
\newblock {\em IEEE Access}, 2024.

\end{thebibliography}

\end{document}